\documentclass[journal]{IEEEtran}

\usepackage{cite}
\usepackage{array}
\usepackage{graphicx}
\usepackage{url}
\usepackage{amsfonts}
\usepackage{multirow}
\usepackage{subcaption}
\usepackage{tabularx}
\usepackage{amsmath}
\usepackage{comment}
\usepackage{array}
\usepackage{amssymb}
\usepackage{import}
\usepackage[leftcaption]{sidecap}
\usepackage{algorithmic}
\usepackage{hyperref}

\DeclareRobustCommand*{\IEEEauthorrefmark}[1]{%
  \raisebox{0pt}[0pt][0pt]{\textsuperscript{\footnotesize #1}}%
}

\begin{document}
\title{2017 Robotic Instrument Segmentation Challenge}
\author{
    \IEEEauthorblockN{
      M. ~Allan\IEEEauthorrefmark{1}, 
      A. ~Shvets\IEEEauthorrefmark{2}, 
      T. ~Kurmann\IEEEauthorrefmark{3}, 
      Z. ~Zhang\IEEEauthorrefmark{4},
      R. ~Duggal\IEEEauthorrefmark{5},
      Y.H. ~Su\IEEEauthorrefmark{6},
      N. ~Rieke\IEEEauthorrefmark{7},
      I. ~Laina\IEEEauthorrefmark{13},
      N. ~Kalavakonda\IEEEauthorrefmark{6},
      S. ~Bodenstedt\IEEEauthorrefmark{8},
      L.C. ~Garcia-Peraza-Herrera\IEEEauthorrefmark{9},
      W. ~Li\IEEEauthorrefmark{15},
      V. ~Iglovikov\IEEEauthorrefmark{10},
      H. ~Luo\IEEEauthorrefmark{11},
      J. ~Yang\IEEEauthorrefmark{12},
      D. ~Stoyanov\IEEEauthorrefmark{9},
      L. ~Maier-Hein\IEEEauthorrefmark{14},
      S. Speidel\IEEEauthorrefmark{8},
      M. ~Azizian\IEEEauthorrefmark{1}\\
    }
    \IEEEauthorblockA{
      \IEEEauthorrefmark{1}Intuitive Surgical Inc., USA,
      \IEEEauthorrefmark{2}Massachusetts Institute of Technology, USA,
      \IEEEauthorrefmark{3}University of Bern, Switzerland,
      \IEEEauthorrefmark{4}University of Alberta, Canada,
      \IEEEauthorrefmark{5}Georgia Institute of Technology, USA,
      \IEEEauthorrefmark{6}University of Washington, USA,
      \IEEEauthorrefmark{7}NVIDIA GmbH., Germany,
      \IEEEauthorrefmark{8}National Center for Tumor Diseases (NCT), Germany,
      \IEEEauthorrefmark{9}Wellcome/EPSRC Centre for Interventional and Surgical Sciences (WEISS) UCL, UK
      \IEEEauthorrefmark{10}Lyft Inc., USA,
      \IEEEauthorrefmark{11}Shenzhen Institute of Advanced Technology, China,
      \IEEEauthorrefmark{12}Beijing Institute of Technology, China,
      \IEEEauthorrefmark{13}Technical University of Munich, Germany,
      \IEEEauthorrefmark{14}German Cancer Research Center, Germany
      \IEEEauthorrefmark{15}Kings College London, UK
  }
}
\maketitle

\begin{abstract}
In mainstream computer vision and machine learning, public datasets such as ImageNet \cite{deng_imagenet_2009}, COCO \cite{lin_coco_2015} and KITTI \cite{geiger_kitti_2012} have helped drive enormous improvements by enabling researchers to understand the strengths and limitations of different algorithms via performance comparison. However, this type of approach has had limited translation to problems in robotic assisted surgery as this field has never established the same level of common datasets and benchmarking methods. 

In 2015 a sub-challenge was introduced at the EndoVis workshop where a set of robotic images were provided with automatically generated annotations from robot forward kinematics. However, there were issues with this dataset due to the limited background variation, lack of complex motion and inaccuracies in the annotation. In this work we present the results of the 2017 challenge on robotic instrument segmentation which involved 10 teams participating in binary, parts and type based segmentation of articulated da Vinci robotic instruments. 

\end{abstract}

\section{Introduction}

\begin{figure}
\centering
\includegraphics[width=0.48\textwidth]{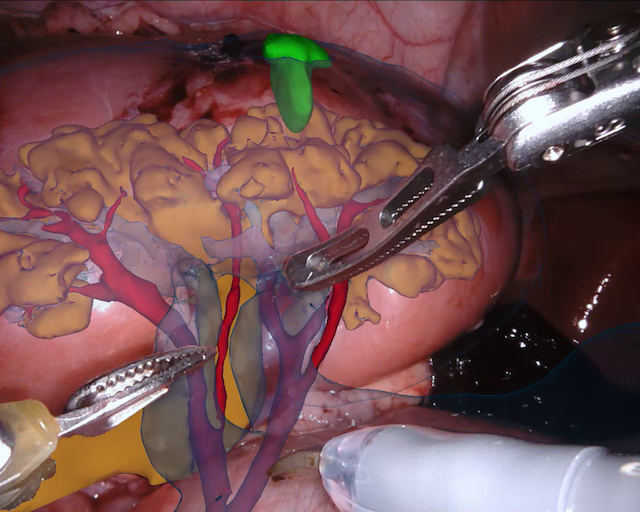}
\caption{\label{fig:augmented_reality_overlay} An example of masking an instrument so that the augmented reality overlay does not occlude the surgeon's view.}
\end{figure}

As robotic minimally invasive surgery has developed, with platforms such as da Vinci$\textregistered$ becoming the de-facto standard-of-care for certain urological, gynecological and general surgical procedures, there has been an increase in focus in how assistive systems based on computer vision and machine learning can improve surgeon performance and patient outcomes. Many potential applications are dependent on scene understanding and for this, accurate segmentation of instruments is an important component. For instance, instrument tracking algorithms which underlie automation and guidance assistance often build upon segmentation \cite{allan_3d_2018} or alternatively masking augmented reality overlays of 3D imaging modalities requires pixel labelling of the instruments to prevent their occlusion (see Fig. \ref{fig:augmented_reality_overlay}). 

Interest in the problem has increased dramatically in the last few years as the dominant methodology has switched from classical machine learning algorithms such as Na{\"i}ve Bayes \cite{lo_episode_2003} and Support Vector Machine (SVM) \cite{bouget_detecting_2015} to deep Convolution Neural Networks (CNNs) \cite{rieke_concurrent_2017,kurmann_simultaneous_2017,pakhomov_deep_2017}. The enormous improvement in performance demonstrated by these techniques has likely been responsible for the increased focus as there begins to be a shift into a position where commercial quality applications become possible.  

However, a significant limitation within the field is the lack of common data and validation sets which allow ease of comparison between methods \cite{bouget_vision_2017} and also prevents further development with possible training of larger networks \cite{pakhomov_deep_2017}. In 2015 the first endoscopic vision sub challenge on instrument segmentation was organized with robotic instrument being one out of four components\footnote{\url{https://endovissub-instrument.grand-challenge.org}}. However, despite heavy usage of this dataset in subsequent publications \cite{rieke_concurrent_2017,kurmann_simultaneous_2017,pakhomov_deep_2017,herrera_toolnet_2018,milletari_cfcm_2018} there are significant limitations due to the limited background variability and also misalignments due to the ground truth being generated by forward kinematics of the da Vinci Research Kit (dVRK) \cite{kazanzides_dvrk_2014} which has significant offsets due to the cable driven joints. 

In 2017 we organized a follow-up challenge\footnote{\url{https://endovissub2017-roboticinstrumentsegmentation.grand-challenge.org}} where a team at Intuitive Surgical manually segmented images from porcine robot assisted nephrectomy procedures. We aimed to improve on the previous challenge by first increasing the label quality by using hand-created labels rather than automatic labelling, secondly by adding greater variance in the background by using 10 separate procedures and finally by providing more type and part labels for the instruments. 

\section{Instrument Segmentation Challenge}

\subsection{Challenge Overview}

Our challenge was made up of 3 sub-problems. The first was binary instrument segmentation, where each frame was separated into da Vinci Xi instruments and a background class, which contained an ultrasound probe, surgical clips and porcine tissues. The second task was instrument part segmentation, where we scored the participants on whether they could correctly segment each articulating part of the instrument (see Fig. \ref{fig:inst_parts}). Our final task was to segment and classify the instruments (see Fig. \ref{fig:inst_types}).

\subsection{Data collection}
\label{subs:data_collection}
Our dataset was made up of 10 sequences of abdominal porcine procedures recorded using da Vinci Xi systems. From each procedure we selected active sequences where significant instrument motion and visibility was observed and sampled 300 frames at a rate of 1 Hz. In cases where instrument motion ceased for several frames we manually removed these frames and extended the sequence so that exactly 300 frames remained. We provided left and right eye images from the stereo camera on the Xi system and also provided camera calibration information in case participants wished to use stereo reconstruction as a feature. 

\begin{figure*}
\centering
\begin{subfigure}[b]{0.50\columnwidth}
\includegraphics[width=\textwidth]{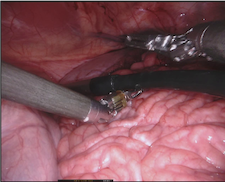}
\end{subfigure}
\hfill
\begin{subfigure}[b]{0.50\columnwidth}
\includegraphics[width=\textwidth]{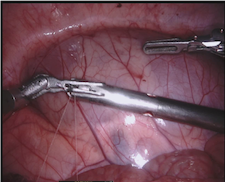}
\end{subfigure}
\hfill
\begin{subfigure}[b]{0.50\columnwidth}
\includegraphics[width=\textwidth]{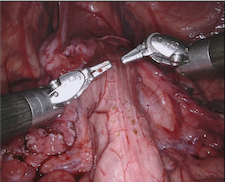}
\end{subfigure}
\hfill
\begin{subfigure}[b]{0.50\columnwidth}
\includegraphics[width=\textwidth]{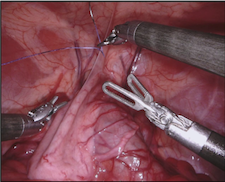}
\end{subfigure}
\\
\vspace{1mm}
\begin{subfigure}[b]{0.50\columnwidth}
\includegraphics[width=\textwidth]{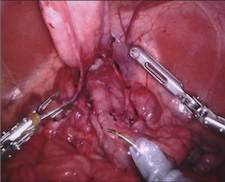}
\end{subfigure}
\hfill
\begin{subfigure}[b]{0.50\columnwidth}
\includegraphics[width=\textwidth]{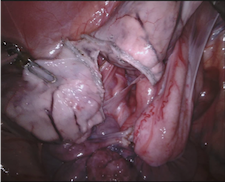}
\end{subfigure}
\hfill
\begin{subfigure}[b]{0.50\columnwidth}
\includegraphics[width=\textwidth]{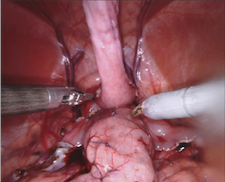}
\end{subfigure}
\hfill
\begin{subfigure}[b]{0.50\columnwidth}
\includegraphics[width=\textwidth]{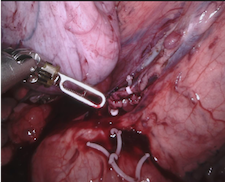}
\end{subfigure}
\caption{\label{fig:training_data} Example frames from the training datasets in order from left to right Dataset 1-8.}
\end{figure*}

\begin{figure}
\includegraphics[width=0.45\textwidth]{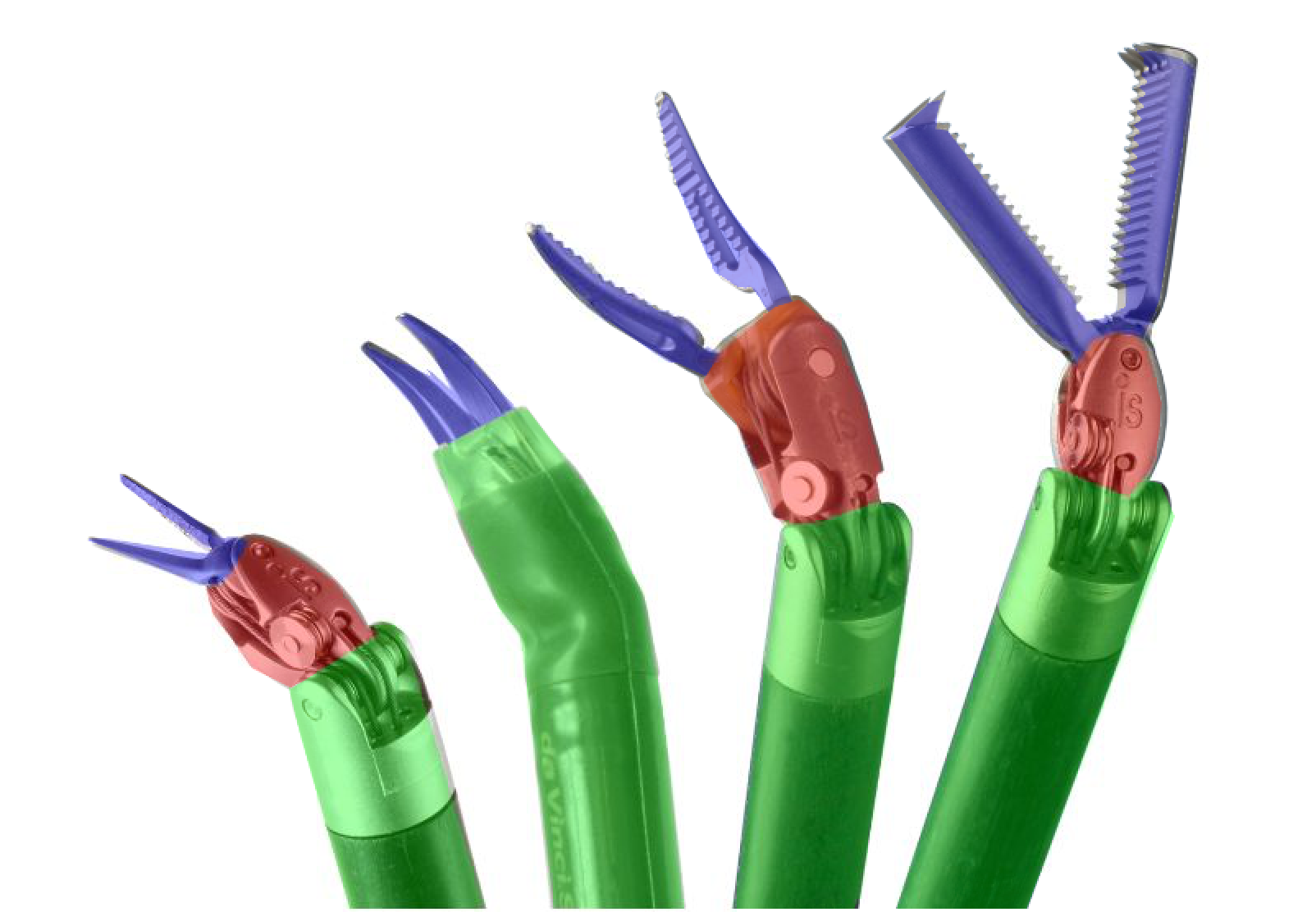}
\caption{\label{fig:inst_parts} A ground truth overlay showing example da Vinci Xi instruments. The different parts of the instrument that are annotate in the parts based segmentation challenge are illustrated with green, red and blue colors. An interesting case is the Monopolar Curved Scissors (2nd from left) which has a protective sheath to insulate the electric current used to provide electro-cautery features. We decided in this case to label the entire sheath as shaft as there is no visible wrist for this instrument.}
\end{figure}

\begin{figure*}
\centering
\begin{subfigure}[b]{0.50\columnwidth}
\includegraphics[width=\textwidth]{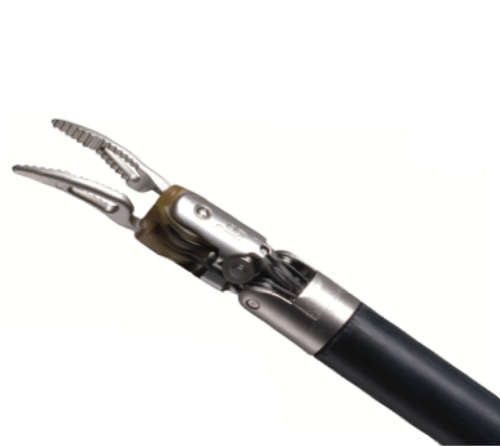}
\caption{}
\end{subfigure}
\hfill
\begin{subfigure}[b]{0.50\columnwidth}
\includegraphics[width=\textwidth]{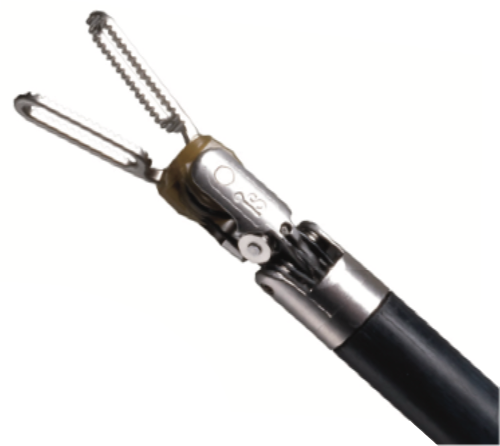}
\caption{}
\end{subfigure}
\hfill
\begin{subfigure}[b]{0.50\columnwidth}
\includegraphics[width=\textwidth]{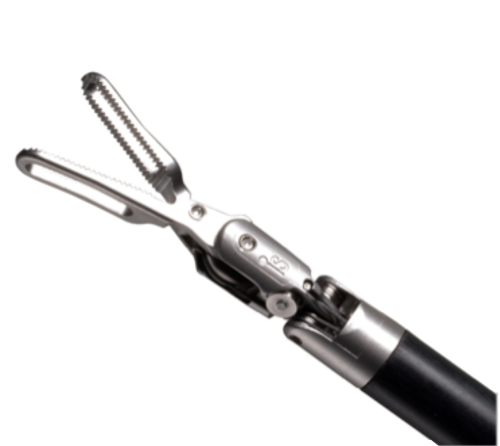}
\caption{}
\end{subfigure}
\hfill
\begin{subfigure}[b]{0.50\columnwidth}
\includegraphics[width=\textwidth]{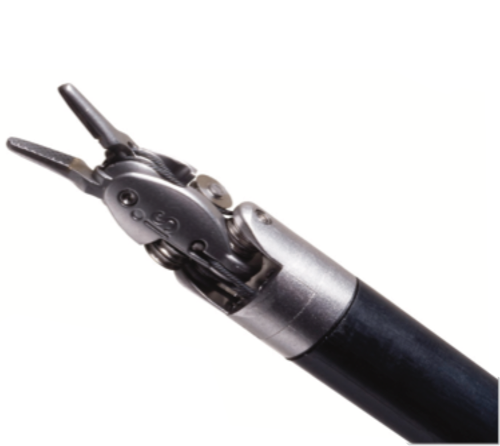}
\caption{}
\end{subfigure}
\\
\begin{subfigure}[b]{0.50\columnwidth}
\includegraphics[width=\textwidth]{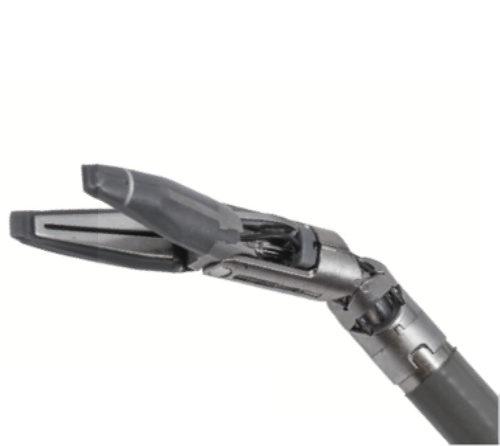}
\caption{}
\end{subfigure}
\hfill
\begin{subfigure}[b]{0.50\columnwidth}
\includegraphics[width=\textwidth]{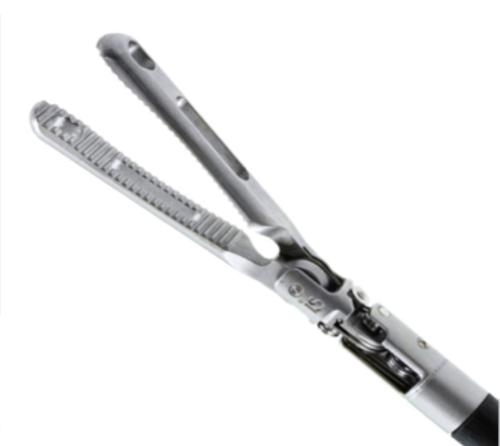}
\caption{}
\end{subfigure}
\hfill
\begin{subfigure}[b]{0.50\columnwidth}
\includegraphics[width=\textwidth]{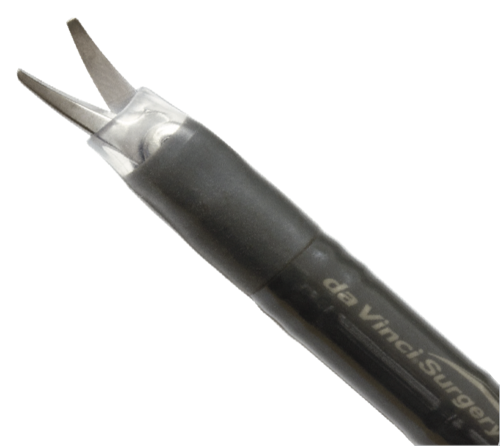}
\caption{}
\end{subfigure}
\hfill
\begin{subfigure}[b]{0.50\columnwidth}
\includegraphics[width=\textwidth]{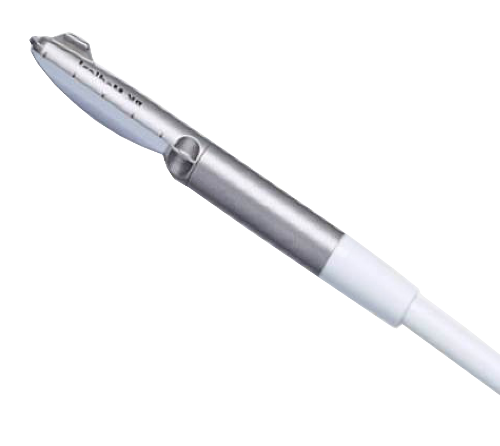}
\caption{}
\end{subfigure}
\caption{\label{fig:inst_types} The different instrument types used in our type based segmentation challenge. (a) shows the Maryland Bipolar Forceps and (b) shows the Fenestrated Bipolar instruments which we combine into a single label Bipolar Forceps due to similar appearance. (c) shows the Prograsp Forceps instrument. (d) shows the Large Needle Driver instrument. (e) shows the Vessel Sealer, the most visually distinctive instrument in our dataset. (f) shows the Grasping Retractor. (h) shows the Monopolar Curved Scissors and (g) shows a drop-in Ultrasound probe from BK Medical which was present in our dataset but not labelled as an instrument.}
\end{figure*}

We provided the first 225 frames of 8 sequences as training data and kept the last 75 frames of those 8 sequences as test data. 2 of the full 300 frame sequences were kept as test sequences. Test labels were kept hidden from the participants. Our datasets contain 7 different robotic surgical instruments. The Large Needle Driver, Prograsp Forceps, Monopolar Curved Scissors, Cadiere Forceps, Bipolar Forceps, Vessel Sealer and additionally a drop-in ultrasound probe, which is typically held in the jaws of the Prograsp Forceps instrument. Samples from the training datasets are depicted in Fig. \ref{fig:training_data} and examples of the different instrument types are shown in Figure \ref{fig:inst_parts} and \ref{fig:inst_types}. 

As we had training sets and test sets from the same surgical sequence, we disallowed participating teams from using the corresponding training set when performing evaluation on each of the 8 split test sets. This resulted in teams needing to train at least 9 models to perform the evaluation. To avoid unfair advantages being given to teams which had access to their own surgical training data, augmenting the dataset with additional data was forbidden. In an exception to this, CNNs pretrained on publicly available, non-surgical data were permitted.

\subsection{Data labelling}

Our labelling was performed by a dedicated segmentation team at Intuitive Surgical using the open source software Viame\footnote{\url{https://github.com/Kitware/VIAME}} which provides functionality for frame-by-frame polygon creation. We labelled only the left eye in the stereo pair to reduce labelling time. Labels were provided on an instance level with separate annotated images per object.

\section{Participating Methods}

\subsection{National Center for Tumor Diseases (Dresden)}

\begin{figure*}[t]
\centering
\includegraphics[width=\textwidth]{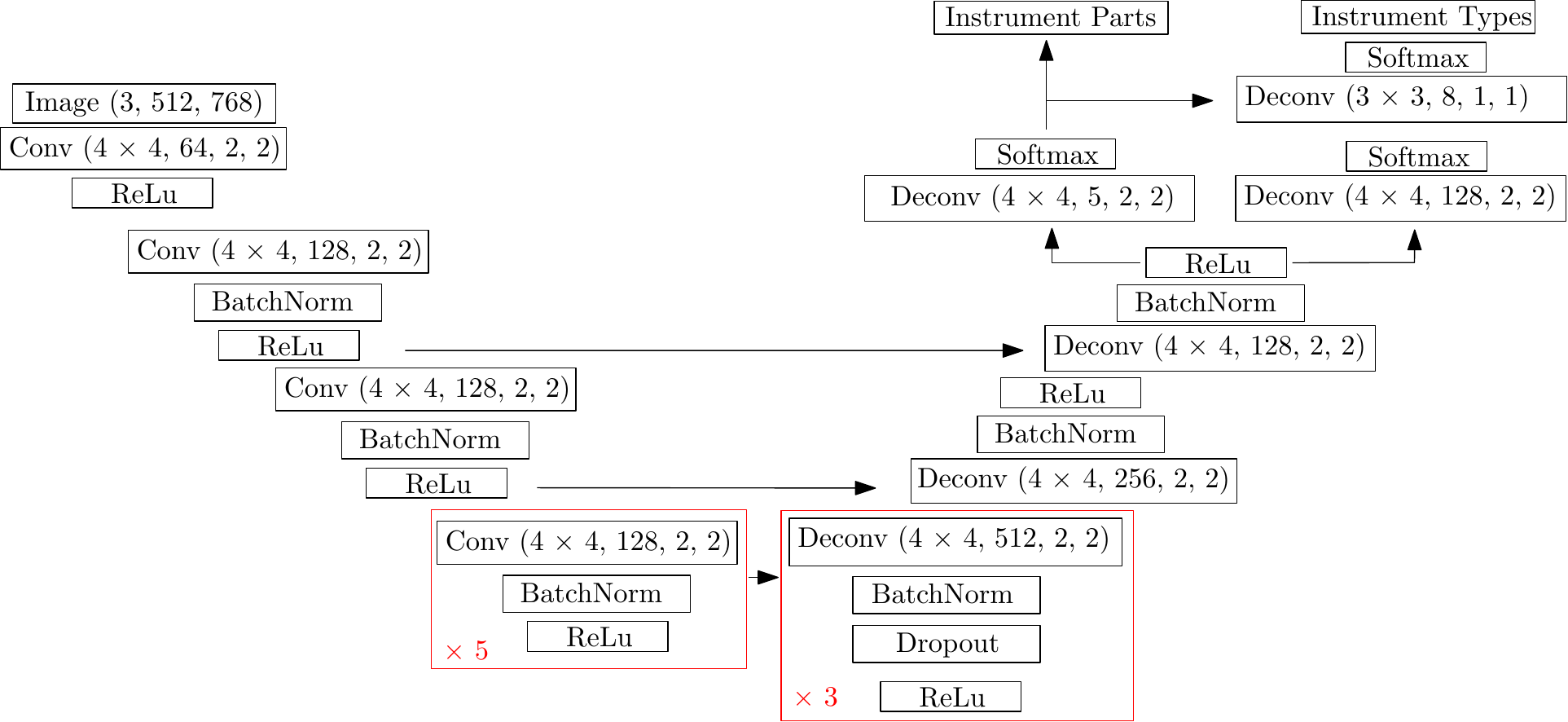}
\caption{\label{fig:method_1}The network architecture from the team at NCT. The convolutional layer notation is kernel size, output dimensions, stride size and padding. The network has two output layers, one providing part-based segmentation and the other providing type segmentation.}
\end{figure*}

Method 1 was from a team at the National Center for Tumor Diseases (NCT) in Dresden. It consisted of Sebastian Bodenstedt, Isabel Funke and Stefanie Speidel. Their method was based on residual CNNs and the topology of the network can be seen in Fig. \ref{fig:method_1}. Before training, they cropped the black borders off the images, equally removing rows and columns to leave an image of resolution $(1536, 1024)$ and then downsampled the image to $(768, 512)$. Random augmentations of increasing the pixel value by $[32,-32]$, vertical and horizontal flips, zooms of $[0.75, 1.25]$ and rotations of $[-10, 10]$ degrees. Training was performed for 200 epochs using a categorical cross entropy as a loss function and Adam as the optimizer. 

\subsection{Universty of Bern}

The method proposed from Thomas Kurmann at the University of Bern (UB) is based on a cascaded Fully Convolution Neural Network (FCN) structure. The cascaded structure attempts to reuse the results of previous stages in order to refine the active stage's result. The first stage of the cascade is the binary segmentation of the instruments. Both instrument part and instrument type segmentation tasks rely on a proper removal of the background, hence profiting from an available binary mask which delineates the background. For all three sub tasks they use the same FCN which is based on an encoder-decoder structure, similar to the U-Net \cite{ronneberger_unet_2015}. Their network uses 6 layers in both the decoder and encoder structure. A layer block is built as follows, Convolution-Batch Normalization (BN)\cite{ioffe2015batch}-Activation-Convolution-BN. They use short skip connections and compute the residuals at the end of each block followed by a ReLu activation. Every encoder block is completed by a max pooling operation with a stride of 2. Decoder layers begin with a transposed convolutional layer to double the size of the input. Before the output they add a last convolution layer with a kernel size of 1x1. All other convolutions in the network use a kernel size of 3x3. The input and output size is 384x384 pixels. Every layer doubles the number of channels in the encoder stage starting with 32. In the decoder stage the number of channels is reduced by 0.5 in every layer. The networks are not pre-trained on any external dataset. The code is available publicly.\footnote{\url{https://github.com/tkurmann/endovis2017_unibe}}

\subsection{Beijing Institute of Technology}

Method 3 was from a team at the Beijing Institute of Technology (BIT). It consisted of Jian Yang, Yakui Chu, Xilin Yang, Zhijia Yang and Shiyun Zhou. To address the problem they trained an ensemble model of 3 separate U-Nets \cite{ronneberger_unet_2015} and then fused the output to obtain a final prediction. 

\subsection{MIT}

Method 4 was from a joint team of Alexey Shvets at MIT and Vladimir Iglovikov at Lyft. As an improvement over vanilla U-Net, they used similar networks with pre-trained encoders. TernausNet is a U-Net-like architecture that uses relatively simple pre-trained VGG11 or VGG16 networks as an encoder \cite{vgg_simonyan_2014, shvets2018automatic}. VGG11 encoder consists of seven convolutional layers, each followed by a ReLU activation function, and five max polling operations, each reducing feature map by 2. The later portion of the network consists of repetitive residual blocks. All convolutional layers have 3x3 kernels. In every residual block, the first convolution operation is implemented with stride 2 to provide down-sampling, while the rest convolution operations use stride 1. In addition, the decoder of the network consists of several decoder blocks that are connected with the corresponding encoder block. As for vanilla U-Net, the transmitted block from the encoder is concatenated to the corresponding decoder block. The output of the model is a pixel-by-pixel mask that shows the class of each pixel. TernausNet16 has a similar structure and uses VGG16 network as an encoder. The code for their approach is available publicly\footnote{\url{https://github.com/ternaus/robot-surgery-segmentation}}.

\subsection{Shenzhen Institute of Advanced Technology}

\begin{figure}
\centering
\includegraphics[width=0.5\textwidth]{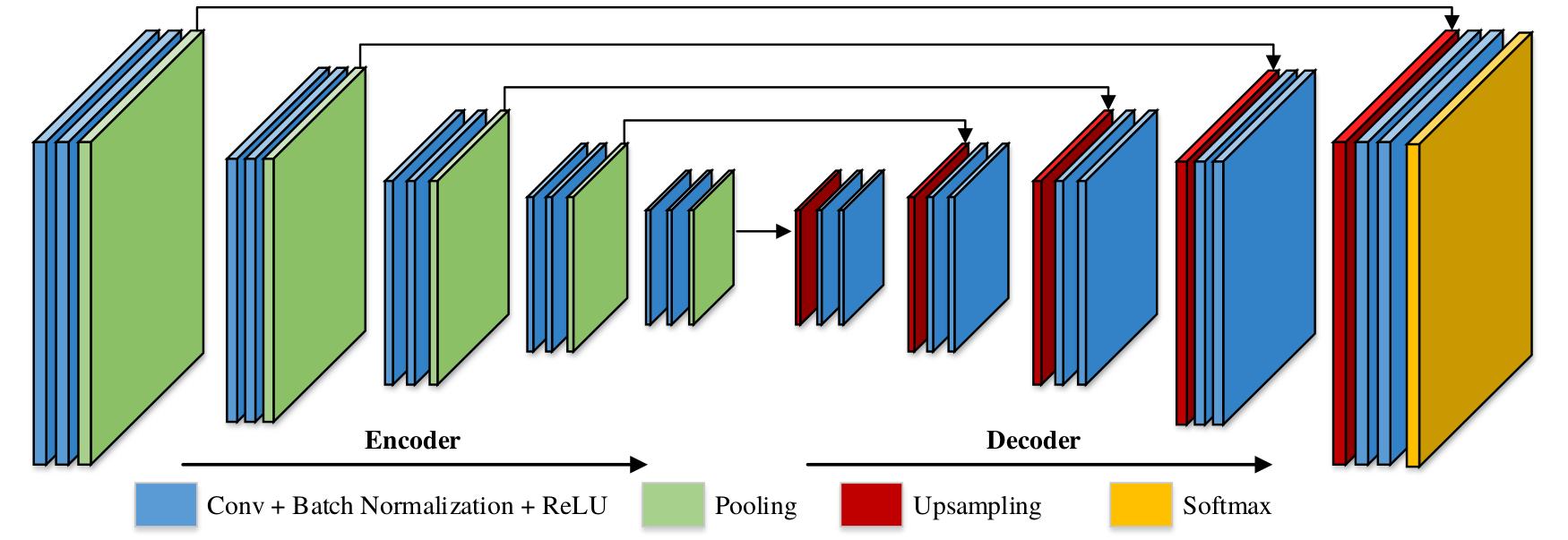}
\caption{\label{fig:shenzen_method.png}The network design from the team from Shenzhen Institute of Advanced Technology.}
\end{figure}

Method 5 was from a team at the Shenzhen Institute of Advanced Technology (SIAT) consisting of Huoling Luo, Ahmed Elazab, Xingguang Duan, Chihua Fang, Qingmao Hu and Fucang Jia. They made use of the SegNet architecture \cite{badrinarayanan_segnet_2015} to address the challenge. This architecture is composed of a symmetric encoder-decoder structure where the output of the decoder is passed into a softmax classification layer (see Fig. \ref{fig:shenzen_method.png}). Following the original paper, a VGG16 encoder-decoder pretrained on ImageNet was used and fine-tuned on the challenge data. Models were trained for the binary and multi-label segmentation tasks, and the multi-label segmentation network is adapted for the type-segmentation task, by taking the weights from the parts segmentation data, modifying the number of outputs to equal the number of classes and fine-tuning this last layer. 


The image aspect ratio was kept unchanged throughout training (via padding) however the images were resized to (480, 270) due to hardware limitations. To handle data imbalance between the instrument wrist and probe classes, a weight compensation strategy was used in the softmax layer:
\begin{equation}
\sigma_{i}(z) = k_{i} \frac{\exp(z_{i})}{\sum_{j=1}^{m} \exp(z_{j})}, i = 1, ..., m
\end{equation}
where $k_{i}$ denotes the class weights, $m$ is the number of instrument part classes and $z_{i}$ is the prediction of class $i$. Data augmentation of random flips and color, contrast and sharpness modifications were applied. Stochastic gradient descent was used to train the models with a learning rate of 0.001 and momentum of 0.9. Training time on an NVIDIA P100 GPU was 12 hours using a batch size of 4. 

\subsection{University College London}

Method 6 was from a team at University College London (UCL) consisting of Luis C. García-Peraza Herrera, Wenqi Li, Tom Vercauteren, and Sebastien Ourselin. They used a custom network design for instrument segmentation called ToolNet \cite{herrera_toolnet_2018} which is inspired by holistically-nested edge detection network \cite{xie_holistically_2015} with an aggregated multiscale loss. Inference runs in real-time. In their original paper they used Dice loss, however in the challenge they modified this to use the intersection-over-union metric:
\begin{equation}
\begin{split}
\mathbf{\hat{y}}^{(\bar{s})}(\mathbf{z}, \boldsymbol{\theta}) &= \sum_{j=1}^{M}w_j \mathbf{\hat{y}}^{(s_j)}(\mathbf{z}, \boldsymbol{\theta}) \\
\mathcal{L}_{MSIoU}(\boldsymbol{y}, \boldsymbol{z}, \boldsymbol{\theta}) &= \bar{\lambda}\mathcal{L}_{IoU}\big(\boldsymbol{\hat{y}^{(\bar{s})}},(\boldsymbol{z}, \boldsymbol{\theta}), \boldsymbol{y}\big) + \\ \sum_{j=1}^{M}\lambda_{j}\mathcal{L}_{IoU}\big(\boldsymbol{\hat{y}^{(s_{j})}},(\boldsymbol{z}, \boldsymbol{\theta}), \boldsymbol{y}\big)
\end{split}
\end{equation}
where $\mathbf{y}$, $\mathbf{z}$ and $\boldsymbol{\theta}$ represent ground truth label, input image, and weights of the network respectively. $\mathbf{\hat{y}}^{(s_j)}(\mathbf{z}, \boldsymbol{\theta})$ represents a probabilistic prediction at scale $j\in\{1, ..., M\}$. $M$ is the number of different scales at which a prediction is generated by the network (i.e. $M=6$ in \texttt{ToolNet}). $\mathbf{\hat{y}}^{(\bar{s})}(\mathbf{z}, \boldsymbol{\theta})$ represents the averaged probabilistic prediction across all scales (i.e. the output of the $1^{2}$ convolution next to all the upsampled predictions in \texttt{ToolNet}). It is worth noting that by learning the weighting parameters $w_j$ (initialized equally for all the scales) for summing predictions coming from multiple scales, we are learning the contribution from each scale to the final loss. $\bar{\lambda}$ and $\lambda_j$ are hyper-parameters that weight the contribution of the losses at different scales (set to $1$ in our implementation).


\begin{figure}[ht]
\centering
\includegraphics[width=0.5\textwidth]{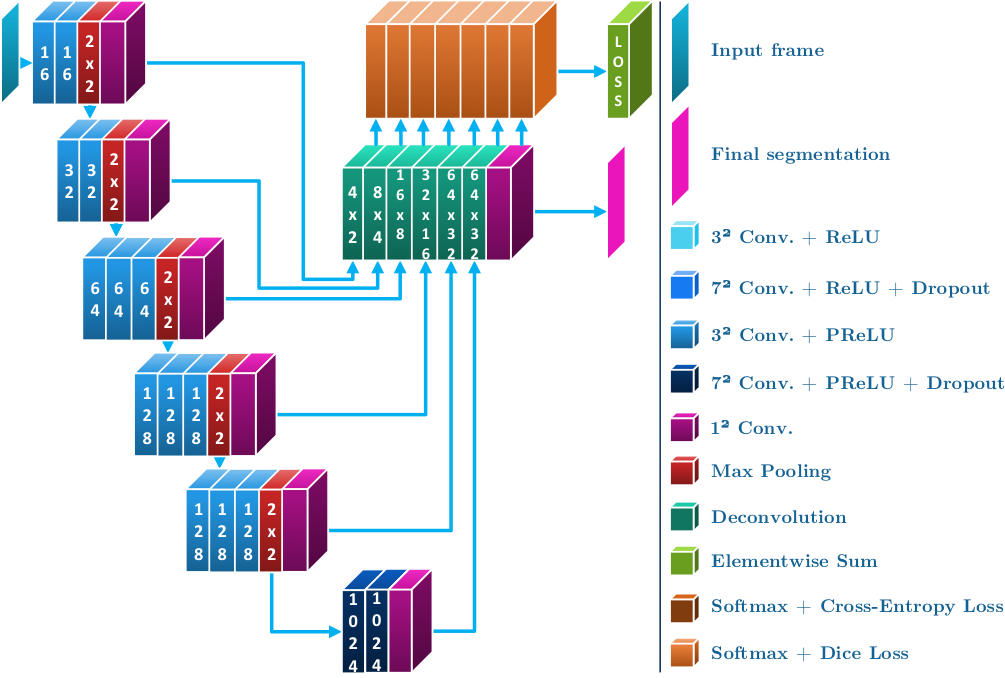}
\caption{\label{fig:ucl}The holistically-nested edge detection network from the UCL team.}
\end{figure}

\subsection{Technical University of Munich}

\begin{figure}[hb]
	\includegraphics[width=0.48\textwidth]{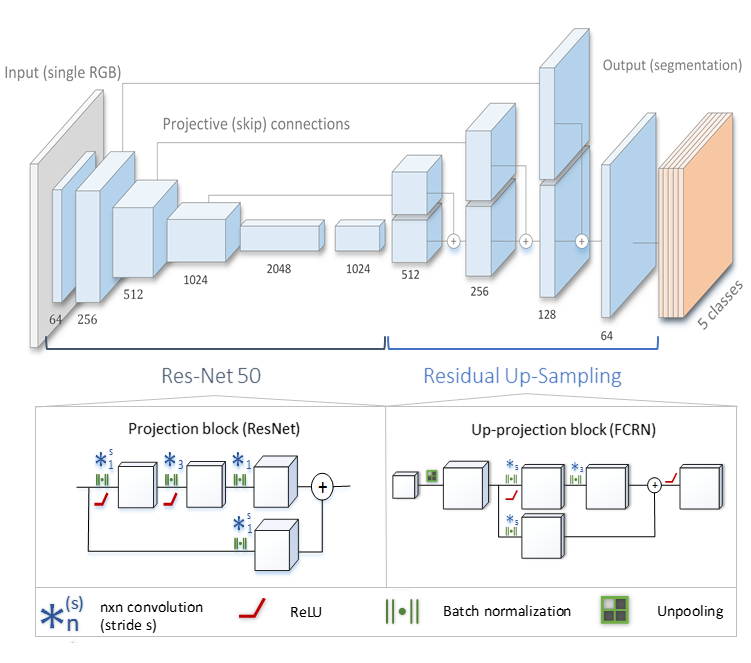}
	\caption{\label{fig:method7}Architecture overview of method 7 from TUM.} 
\end{figure}

Method 7 was from a team at the Technical University of Munich (TUM) consisting of Nicola Rieke and Iro Laina. 
The presented method is based on the $CSL$ network for simultaneous segmentation and 2D pose estimation of the instrument that they published with joint first authorship in the main conference \cite{rieke_concurrent_2017}. 
As the challenge data does not provide points of interest to track, the network architecture is reduced to the segmentation-only version of $CSL$ method (see Fig. \ref{fig:method7}). 

The network follows an encoder-decoder structure based on a Fully Convolutional Residual Network (FCRN) \cite{laina2016deeper} with added long-range skip connections. 
ResNet-50 \cite{he2016deep} is used as the encoder, which maps the input frames of resolution $(480, 480)$ to feature maps of lower resolution through a set of residual blocks. The decoder consists of residual up-sampling layers which successively increase the feature map resolution to the output space where they set one output channel per class.
To counteract the loss of spatial information due to downsampling, skip connections are added to allow the gradient to bypass part of the network and flow directly from encoding layers to decoding layers. 
These long-range skip connections reshape the respective encoding layer with a 1x1 convolution and add it to the decoding layer of the same resolution. 
Separate models are trained for binary segmentation (2 classes) and part segmentation (5 classes).

During training, standard photometric and geometric augmentations are employed to extend the variability of the training dataset. 
In addition, an application-specific augmentation is introduced to increase the robustness of the proposed model against specular reflections on the instruments which are often the cause of misclassification. Thus, as a form of augmentation, specularities of random strength and size are added along the shaft of the instruments. 

In contrast to the original publication, an additional post-processing operation is performed to reduce the noise.
Due to the surgical setup, the instruments always enter the recorded scene from one of the image borders. This prior knowledge is included indirectly as a post-processing step by computing the connected components and assigning the background class to spurious instrument predictions that are not connected to the border. Morphological operations are applied to fill holes and make the prediction smoother. 

The resulting method runs with near real-time performance on a NVIDIA GeForce GTX TITAN X.

\subsection{IIIT Delhi}
Method 8 was from the Indraprastha institute of Information Technology in Delhi and consisted of Rahul Duggal and Dr Anubha Gupta. Their approach was geared towards answering the question - how well does a simplistic baseline perform? Essentially, their deep learning based approach consisted of a CNN trained as a foreground detector followed by a Conditional Random Field (CRF) based post processing \cite{crf_krahenbuhl_2011}. The foreground detector was a VGG-19 \cite{vgg_simonyan_2014} model trained on $51 \times 51$ patches from the original image, with ground patched containing any portion of an instrument labelled $1$, and others $0$. The code for their approach is available publicly\footnote{\url{https://github.com/duggalrahul/MICCAI17_EndoVis_RoboSeg}}.

\subsection{University of Alberta}

\begin{figure}[tbph]
\includegraphics[width=0.48\textwidth]{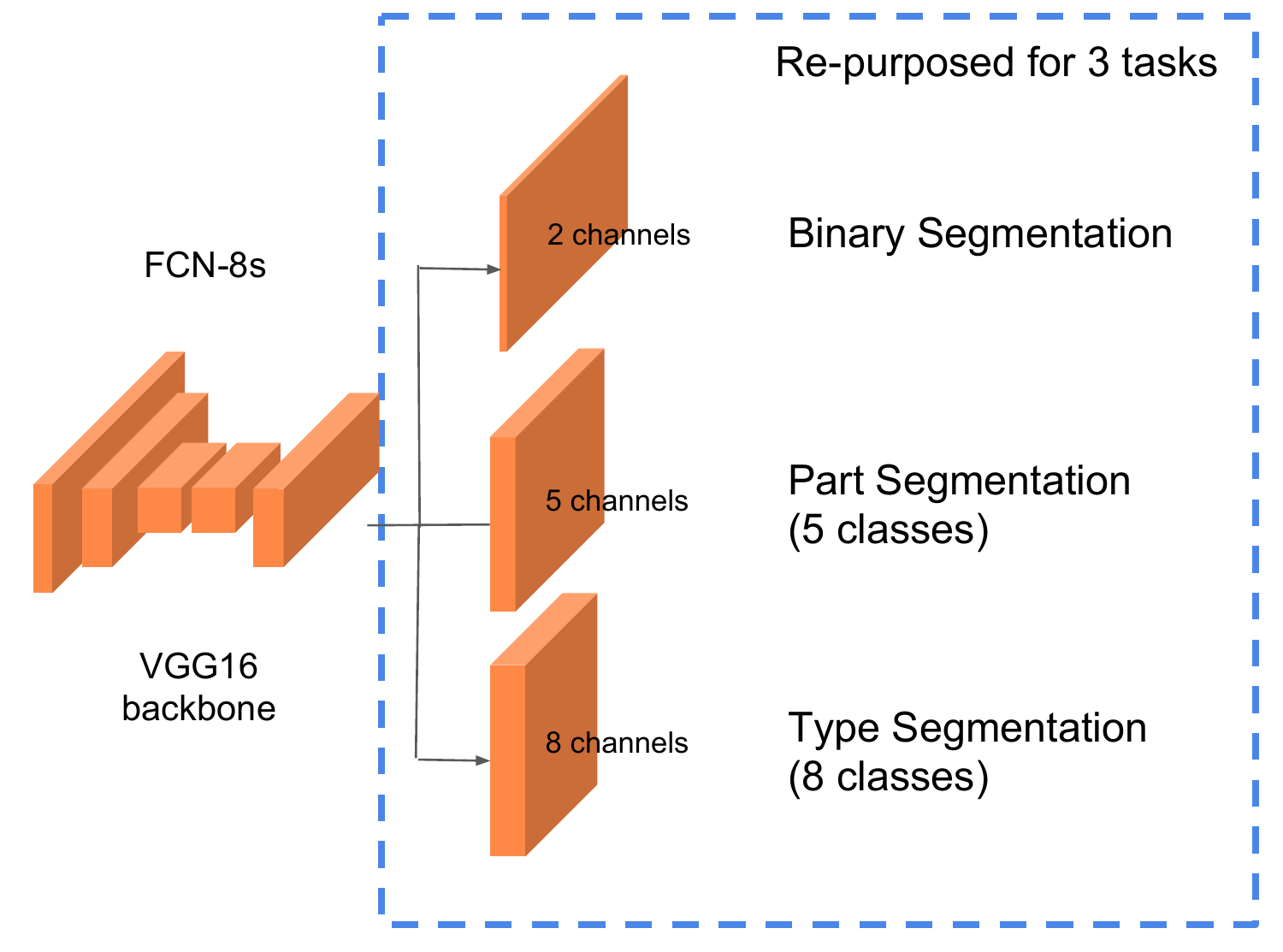}
\caption{\label{fig:method8-arch}Architecture overview of the submission from University of Alberta} 
\end{figure}

Method 9 was from the University of Alberta (UA). The team consisted of members Zichen (Vincent) Zhang, Xuebin Qin, Min Tang, Shida He, Dana Cobzas and Martin Jagersand.
Their method was based on FCN \cite{long_fully_2015} and the overall architecture is illustrated in Fig. \ref{fig:method8-arch}. It consists of a repurposed FCN-8s for all three tasks. This network architecture was used since it's relatively smaller than other popular choices such as Resnet-101 \cite{he2016deep}. For each video, there are only 1575 (225 $\times$ 7) frames (not considering data augmentation) for training the network. A smaller network would help reduce the risk of overfitting in this setting.

For datasets 1-8, they trained a separate model for each. That is, for each dataset, training was performed using the other seven videos and validation was done on the target video. For dataset 9 and 10, since the entire videos were in the test set, there was no need for excluding any training data. Due to time constraints, they had to reuse the model trained from the dataset 1-8. They ended up picking the model trained for dataset 7 since this dataset contained data that were rarely seen in the other datasets and was perhaps the least useful to train the network on (note that this model was trained on dataset 1-6 and 8).
In total, they trained 8 sets of weights for each task, i.e. 24 models in total. The hyperparameters were fixed for all datasets and all tasks. In the binary segmentation task, the network weights were initialized with the pretrained model on PASCAL dataset \cite{long_fully_2015} and fine-tuned on it. The best weights for this task were then used to initialize the network in the part and type segmentation tasks. The input images were resized to 320 $\times$ 256 for faster training.

The network prediction contained some small false positive regions. In post-processing, connected regions smaller than 15000 pixels were removed. This threshold was tuned on the training data.

\subsection{University of Washington}

Method 10 was from the University of Washington (UW) and the team was made up of Yun-Hsuan Su and Niveditha Kalavakonda.
Unlike the solutions from many other teams, they are interested in developing a surgical tool segmentation method without machine learning, and see how far traditional computer vision approaches can go in this matter. The motivation comes from the lack of massive pre-labeled surgical image dataset in general situations \cite{kohli2017medical}. In fact, this work is inspired by another study by the team - surgical tool segmentation with robot kinematics prior \cite{su2018real}. Few advantages for this algorithm are that no training is required, and efficient on-line execution is possible without GPU. 

Color filtering was used to initially generate a mask of instruments to distinguish from the background, where color features are extracted using thresholds on the opponent color space and hue/saturation channels. The initial mask was then fed to the Grabcut Algorithm \cite{rother2004grabcut} for refining the prediction of tool versus tissue. Many image features including shape masking, edge constraints, border constraints, and disparity discontinuity have potential impact on the segmentation result. Depending on the blurriness of the image, the weights for the chosen image features vary. For example, edge constraints and disparity information are less reliable in blurry images, and can be misleading in the case of interlacing \cite{soper2008mastery}. To address this, a high level classification of image blurriness is performed using the variance of the response of a Laplacian kernel. Upon computing a weighted sum of the features, one may be able to determine whether each pixel belongs to a tool or tissue based on blurriness score. Finally, a probability mask is generated and the border constraint prior is enforced to remove erroneous islands from the final mask. This is determined with the knowledge that contours for instruments will be connected to at least one of the four edges of the image.

\section{Results}

\subsection{Evaluation Criteria}

Our evaluation criteria for each challenge was based on mean intersection-over-union (IoU), a current standard for assessing segmentation scores in computer vision literature \cite{lin_coco_2015}. The IoU for a single class is defined as 
\begin{equation}
IOU = TP/(TP+FP+FN)
\end{equation}
where TP is the number of true positive predictions for a class label, FP is the number of false positives and FN is the number of false negatives. To compute the mean IoU we use the arithmetic mean of the IoU for all classes that are present in a given frame. If we are considering a set of classes and none are present in the frame, we discount the frame from the calculation. We compute this score for each frame and average over all frames to get a per-dataset score. When computing overall scores we weight each score by the size of the dataset. 

Rather than releasing test labels directly to teams, the annotations were kept private and teams made a single submission of their segmentations. On the challenge day at MICCAI 2017 the results were made public. This meant that teams were not able to tune their methods on the test data by making multiple submissions. However, as the entire training set was released prior to the challenge day and we had to rely on the fairness of the teams to follow the challenge rules to exclude the corresponding training set for a test set, as described in Section \ref{subs:data_collection}.

\subsection{Binary Segmentation}

Our first challenge was binary segmentation, where the objective was to divide images into a class made up of any da Vinci surgical instrument and any other object, anatomical or man-made. Although this challenge was by far the simplest of the three, there were significant challenges due to drop-in US probes and needles having quite similar appearance and color to surgical instruments. The numerical results of computing the mean IoU for each dataset for each of the 10 participating teams are shown in Table \ref{fig:binary_numerical_results}. The highest scores for each dataset were shared between the methods of University of Bern (UB) and MIT and the highest average score was MIT with 0.854. In total 5 teams scored an average of over 0.8 for mean IoU over all datasets. The worst scoring dataset was dataset 1 with a mean IoU of 0.589 yet several methods (UB and MIT) were able to score above 0.8 for this dataset. 

\begin{figure*}
\captionsetup[subfigure]{labelformat=empty}
\begin{subfigure}[b]{0.23\textwidth}
\includegraphics[width=\textwidth]{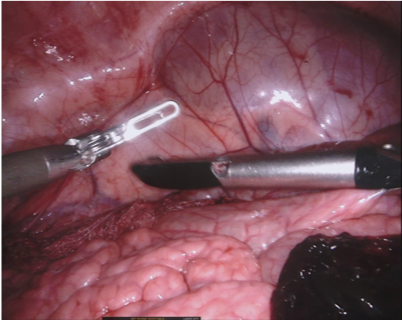}
\caption{Frame 278}
\end{subfigure}
\hfill
\begin{subfigure}[b]{0.23\textwidth}
\includegraphics[width=\textwidth]{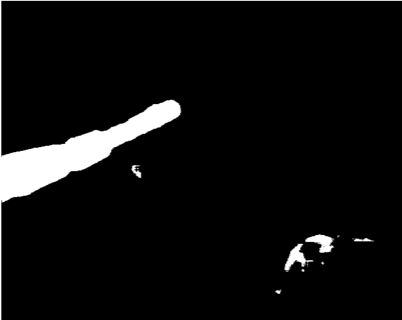}
\caption{NCT}
\end{subfigure}
\hfill
\begin{subfigure}[b]{0.23\textwidth}
\includegraphics[width=\textwidth]{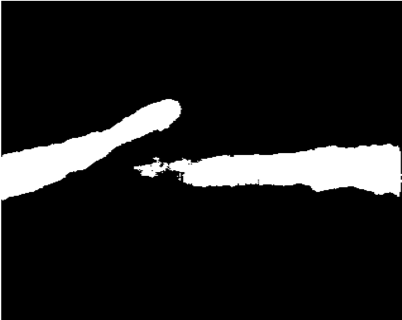}
\caption{SIAT}
\end{subfigure}
\hfill
\begin{subfigure}[b]{0.23\textwidth}
\includegraphics[width=\textwidth]{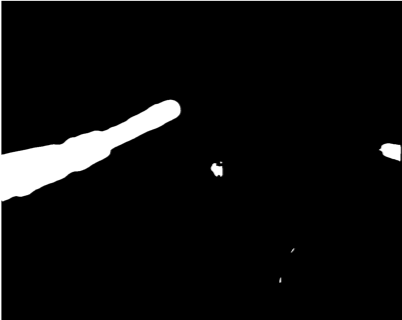}
\caption{MIT}
\end{subfigure}
\\
\begin{subfigure}[b]{0.23\textwidth}
\includegraphics[width=\textwidth]{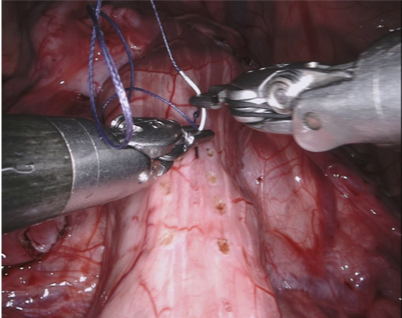}
\caption{Frame 237}
\end{subfigure}
\hfill
\begin{subfigure}[b]{0.23\textwidth}
\includegraphics[width=\textwidth]{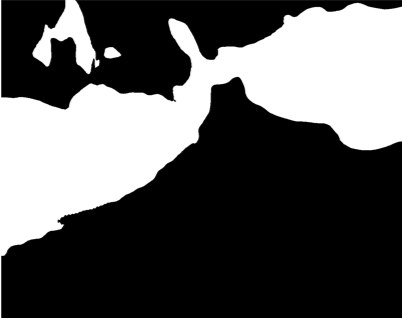}
\caption{BIT}
\end{subfigure}
\hfill
\begin{subfigure}[b]{0.23\textwidth}
\includegraphics[width=\textwidth]{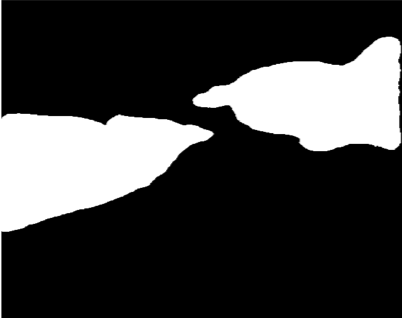}
\caption{TUM}
\end{subfigure}
\hfill
\begin{subfigure}[b]{0.23\textwidth}
\includegraphics[width=\textwidth]{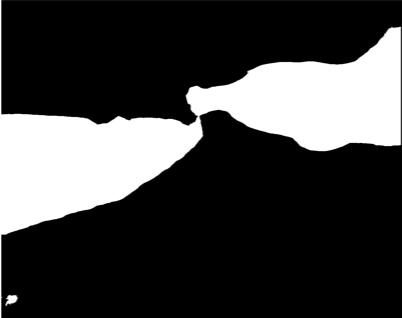}
\caption{UCL}
\end{subfigure}
\begin{subfigure}[b]{0.23\textwidth}
\includegraphics[width=\textwidth]{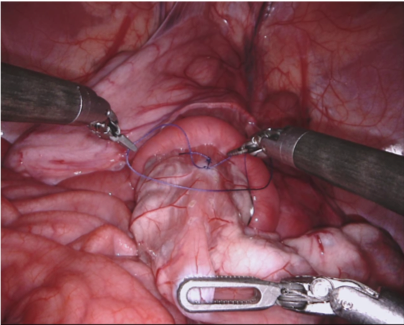}
\caption{Frame 251}
\end{subfigure}
\hfill
\begin{subfigure}[b]{0.23\textwidth}
\includegraphics[width=\textwidth]{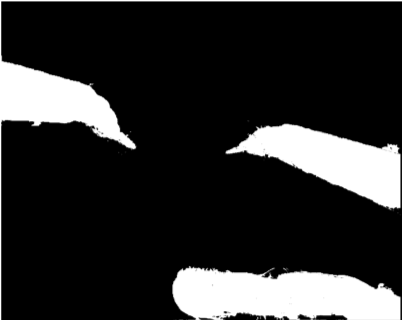}
\caption{IIIT Delhi}
\end{subfigure}
\hfill
\begin{subfigure}[b]{0.23\textwidth}
\includegraphics[width=\textwidth]{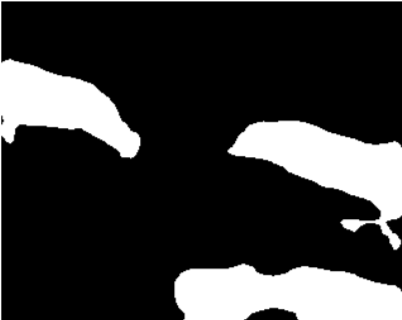}
\caption{U Alberta}
\end{subfigure}
\hfill
\begin{subfigure}[b]{0.23\textwidth}
\includegraphics[width=\textwidth]{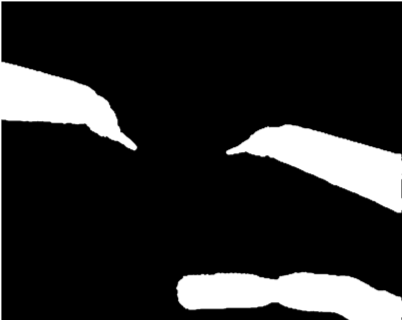}
\caption{U Bern}
\end{subfigure}
\\
\begin{subfigure}[b]{0.23\textwidth}
\includegraphics[width=\textwidth]{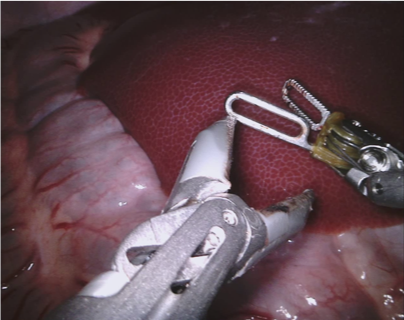}
\caption{Frame 247}
\end{subfigure}
\hfill
\begin{subfigure}[b]{0.23\textwidth}
\includegraphics[width=\textwidth]{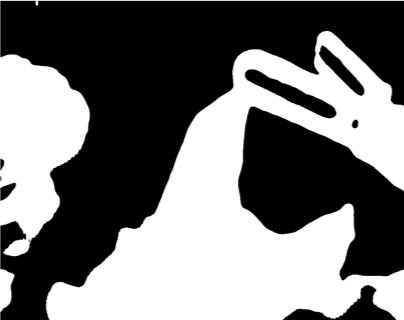}
\caption{BIT}
\end{subfigure}
\hfill
\begin{subfigure}[b]{0.23\textwidth}
\includegraphics[width=\textwidth]{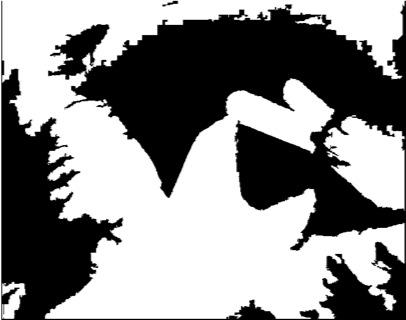}
\caption{U Washington}
\end{subfigure}
\hfill
\begin{subfigure}[b]{0.23\textwidth}
\includegraphics[width=\textwidth]{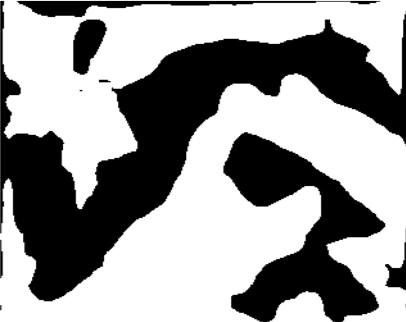}
\caption{U Alberta}
\end{subfigure}
\hfill
\begin{subfigure}[b]{0.23\textwidth}
\includegraphics[width=\textwidth]{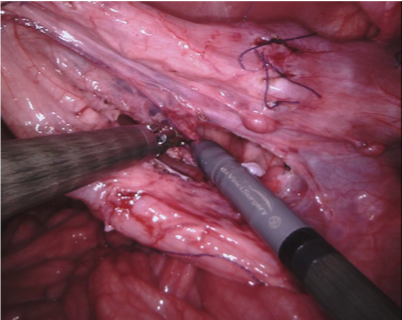}
\caption{Frame 256}
\end{subfigure}
\hfill
\begin{subfigure}[b]{0.23\textwidth}
\includegraphics[width=\textwidth]{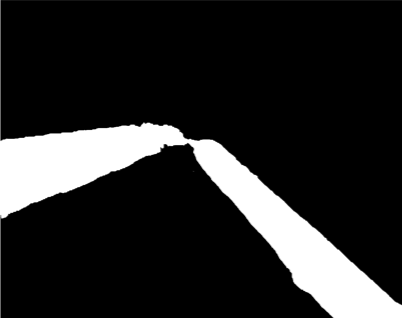}
\caption{U Bern}
\end{subfigure}
\hfill
\begin{subfigure}[b]{0.23\textwidth}
\includegraphics[width=\textwidth]{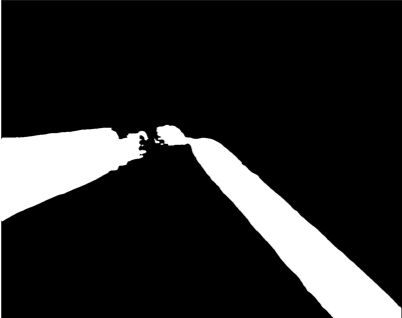}
\caption{MIT}
\end{subfigure}
\hfill
\begin{subfigure}[b]{0.23\textwidth}
\includegraphics[width=\textwidth]{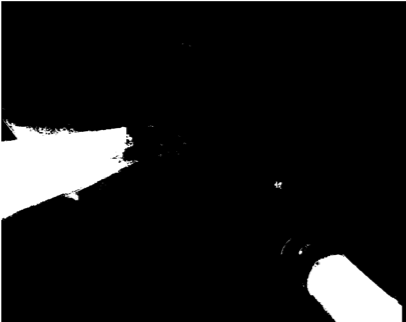}
\caption{IIIT Delhi}
\end{subfigure}
\hfill
\begin{subfigure}[b]{0.23\textwidth}
\includegraphics[width=\textwidth]{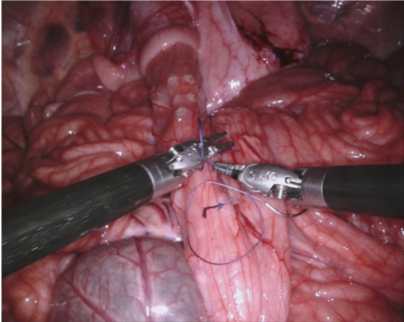}
\caption{Frame 101}
\end{subfigure}
\hfill
\begin{subfigure}[b]{0.23\textwidth}
\includegraphics[width=\textwidth]{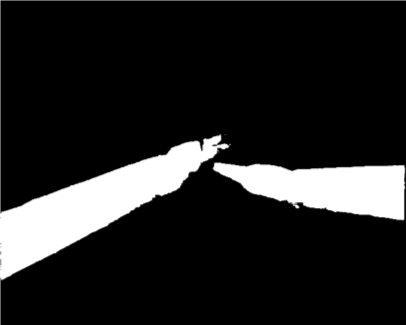}
\caption{NCT}
\end{subfigure}
\hfill
\begin{subfigure}[b]{0.23\textwidth}
\includegraphics[width=\textwidth]{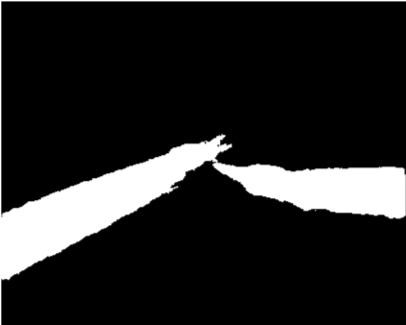}
\caption{SIAT}
\end{subfigure}
\hfill
\begin{subfigure}[b]{0.23\textwidth}
\includegraphics[width=\textwidth]{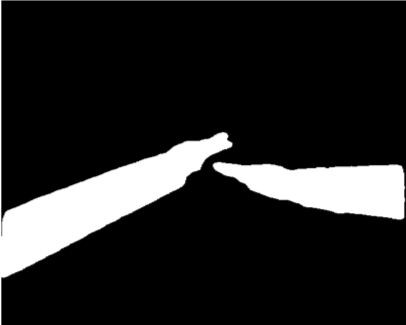}
\caption{TUM}
\end{subfigure}
\caption{\label{fig:binary_segmentation_qualitative}Qualitative results showing frames from different datasets with the corresponding results from randomly selected methods. In order, the datasets shown are Dataset 1, Dataset 3, Dataset 4, Dataset 7, Dataset 8 and Dataset 10.}
\end{figure*}

In Fig. \ref{fig:binary_segmentation_qualitative} we show qualitative results of randomly chosen frames from each dataset. In the top row, we show frame 278 from dataset 1, which contained 2 Prograsp Forceps instruments and a drop in ultrasound probe. The 3 selected frames were from methods that all averaged over 0.8 mean IoU yet showed considerable difference in their ability to differentiate the US probe. There was also visibly different performance across the methods in dataset 7, which contained a Vessel Sealer and complex lighting.

\begin{table*}
\centering
\small
\begin{tabular}{ l |  c | c | c | c | c | c | c | c | c | c | c }
 & NCT & UB & BIT & MIT & SIAT & UCL & TUM & Delhi & UA & UW & Mean IoU \\
\hline
Dataset 1  & 0.784  & 0.807  & 0.275  & \textbf{0.854}  & 0.625  & 0.631  & 0.760  & 0.408  & 0.413  & 0.337 & 0.589 \\
Dataset 2  & 0.788  & \textbf{0.806}  & 0.282  & 0.794  & 0.669  & 0.645  & 0.799  & 0.524  & 0.463  & 0.289 & 0.606 \\
Dataset 3  & 0.926  & 0.914  & 0.455  & \textbf{0.949}  & 0.897  & 0.895  & 0.916  & 0.743  & 0.703  & 0.483 & 0.788 \\
Dataset 4  & 0.934  & 0.925  & 0.310  & \textbf{0.949}  & 0.907  & 0.883  & 0.915  & 0.782  & 0.751  & 0.678 & 0.803 \\
Dataset 5  & 0.701  & 0.740  & 0.220  & \textbf{0.862}  & 0.604  & 0.719  & 0.810  & 0.528  & 0.375  & 0.219 & 0.578 \\
Dataset 6  & 0.876  & 0.890  & 0.338  & \textbf{0.922}  & 0.843  & 0.852  & 0.873  & 0.292  & 0.667  & 0.619 & 0.717 \\
Dataset 7  & 0.846  & \textbf{0.930}  & 0.404  & 0.856  & 0.832  & 0.710  & 0.844  & 0.593  & 0.362  & 0.325 & 0.670 \\
Dataset 8  & 0.881  & 0.904  & 0.366  & \textbf{0.937}  & 0.513  & 0.517  & 0.895  & 0.562  & 0.797  & 0.506 & 0.688 \\
Dataset 9  & 0.789  & 0.855  & 0.236  & 0.865  & 0.839  & 0.808  & \textbf{0.877}  & 0.626  & 0.539  & 0.377 & 0.681 \\
Dataset 10  & 0.899  & \textbf{0.917}  & 0.403  & 0.905  & 0.899  & 0.869  & 0.909  & 0.715  & 0.689  & 0.603 & 0.781 \\
\hline
Mean IOU & 0.843  & 0.875  & 0.326  & \textbf{0.888}  & 0.803  & 0.785  & 0.873  & 0.612  & 0.591  & 0.461
\end{tabular}
\caption{\label{fig:binary_numerical_results} The numerical results for the binary segmentation task. The highest scoring method is shown in bold. 6 Datasets were won by the team from MIT, 3 by the team from UB and 1 by the team from TUM.}
\end{table*}

\subsection{Parts Segmentation}

Our second challenge was on instrument part segmentation where the participants were challenged to divide the binary instrument labels into a shaft, wrist and jaws. As in the binary segmentation challenge, the drop-in US probe and other man-made devices as well as all anatomical objects were to be labelled as background. We compute the mean IoU for each frame of each dataset and for frames where no instance of a class occurred, such as when the shaft is withdrawn completely from the field of view. Nine teams participated in this challenge, the only team abstaining was from IIT Delhi.

In Table \ref{fig:parts_numerical_results} we show quantitative results for the 9 participating teams. The TUM team achieved the highest overall accuracy achieving a mean IoU of 0.751 however the MIT achieved the highest mean IoU in 7 of the 10 datasets. The TUM score averaged higher due to the larger weighting on Datasets 9 and 10 which contained 4x as many frames as Datasets 1-8.

\begin{figure*}
\captionsetup[subfigure]{labelformat=empty}
\begin{subfigure}[b]{0.23\textwidth}
\includegraphics[width=\textwidth]{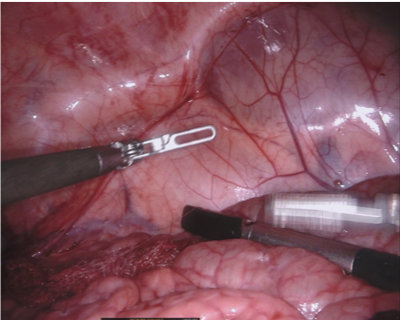}
\caption{Frame 278}
\end{subfigure}
\hfill
\begin{subfigure}[b]{0.23\textwidth}
\includegraphics[width=\textwidth]{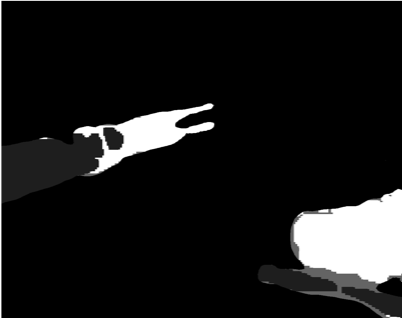}
\caption{BIT}
\end{subfigure}
\hfill
\begin{subfigure}[b]{0.23\textwidth}
\includegraphics[width=\textwidth]{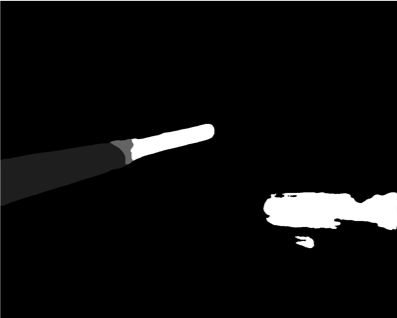}
\caption{MIT}
\end{subfigure}
\hfill
\begin{subfigure}[b]{0.23\textwidth}
\includegraphics[width=\textwidth]{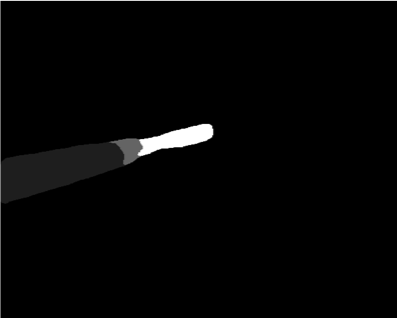}
\caption{TUM}
\end{subfigure}
\\
\begin{subfigure}[b]{0.23\textwidth}
\includegraphics[width=\textwidth]{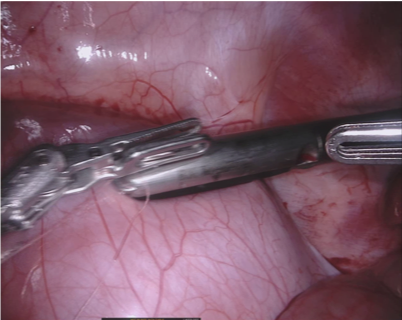}
\caption{Frame 227}
\end{subfigure}
\hfill
\begin{subfigure}[b]{0.23\textwidth}
\includegraphics[width=\textwidth]{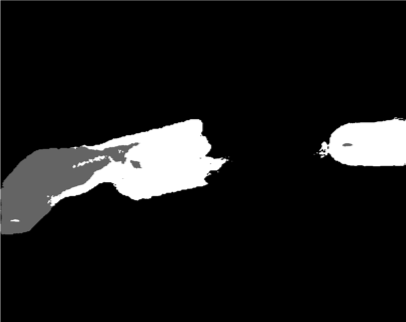}
\caption{U Bern}
\end{subfigure}
\hfill
\begin{subfigure}[b]{0.23\textwidth}
\includegraphics[width=\textwidth]{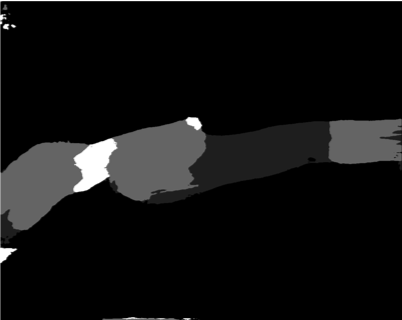}
\caption{UCL}
\end{subfigure}
\hfill
\begin{subfigure}[b]{0.23\textwidth}
\includegraphics[width=\textwidth]{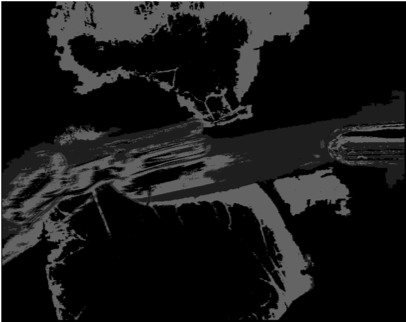}
\caption{U Washington}
\end{subfigure}
\\
\begin{subfigure}[b]{0.23\textwidth}
\includegraphics[width=\textwidth]{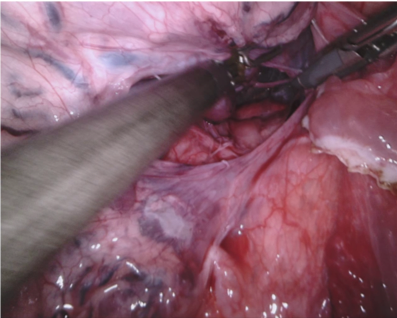}
\caption{Frame 257}
\end{subfigure}
\hfill
\begin{subfigure}[b]{0.23\textwidth}
\includegraphics[width=\textwidth]{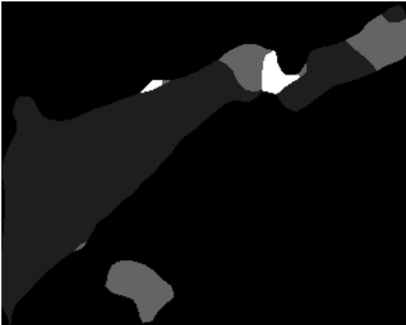}
\caption{U Alberta}
\end{subfigure}
\hfill
\begin{subfigure}[b]{0.23\textwidth}
\includegraphics[width=\textwidth]{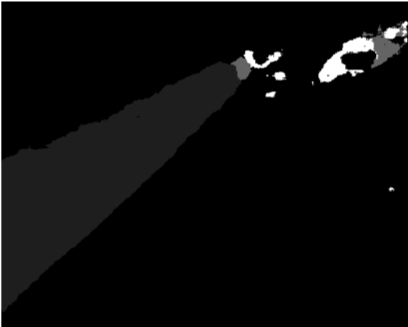}
\caption{NCT}
\end{subfigure}
\hfill
\begin{subfigure}[b]{0.23\textwidth}
\includegraphics[width=\textwidth]{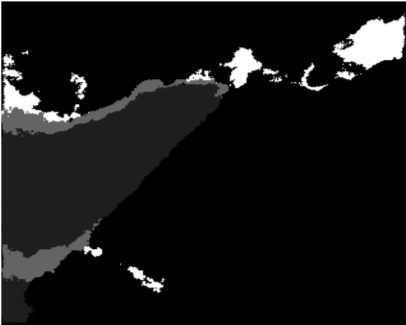}
\caption{SIAT}
\end{subfigure}
\\
\begin{subfigure}[b]{0.23\textwidth}
\includegraphics[width=\textwidth]{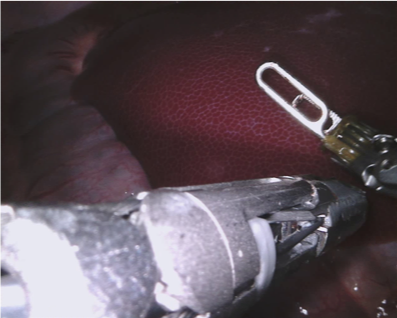}
\caption{Frame 230}
\end{subfigure}
\hfill
\begin{subfigure}[b]{0.23\textwidth}
\includegraphics[width=\textwidth]{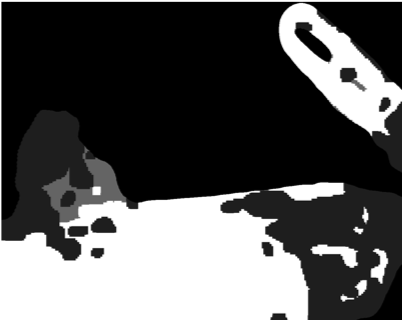}
\caption{BIT}
\end{subfigure}
\hfill
\begin{subfigure}[b]{0.23\textwidth}
\includegraphics[width=\textwidth]{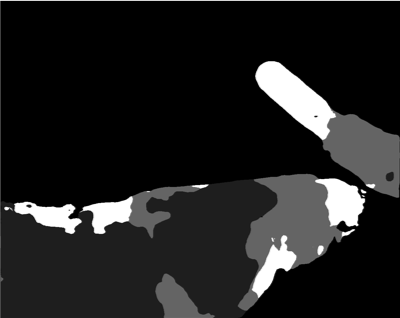}
\caption{MIT}
\end{subfigure}
\hfill
\begin{subfigure}[b]{0.23\textwidth}
\includegraphics[width=\textwidth]{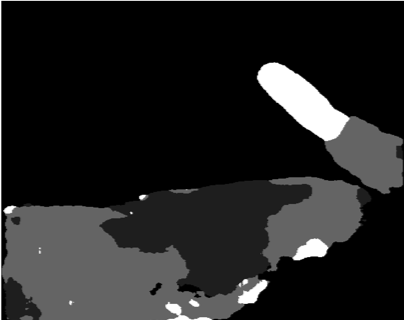}
\caption{TUM}
\end{subfigure}
\\
\begin{subfigure}[b]{0.23\textwidth}
\includegraphics[width=\textwidth]{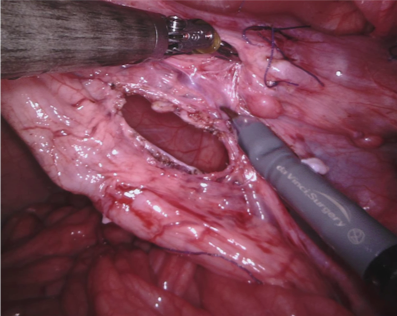}
\caption{Frame 224}
\end{subfigure}
\hfill
\begin{subfigure}[b]{0.23\textwidth}
\includegraphics[width=\textwidth]{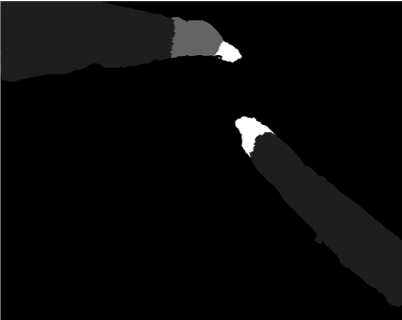}
\caption{U Bern}
\end{subfigure}
\hfill
\begin{subfigure}[b]{0.23\textwidth}
\includegraphics[width=\textwidth]{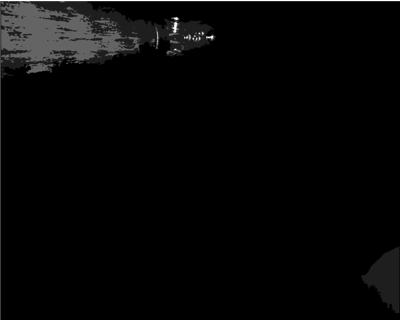}
\caption{U Washington}
\end{subfigure}
\hfill
\begin{subfigure}[b]{0.23\textwidth}
\includegraphics[width=\textwidth]{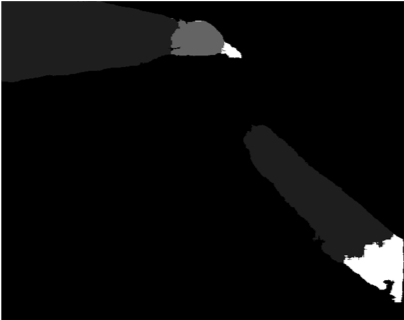}
\caption{NCT}
\end{subfigure}
\\
\begin{subfigure}[b]{0.23\textwidth}
\includegraphics[width=\textwidth]{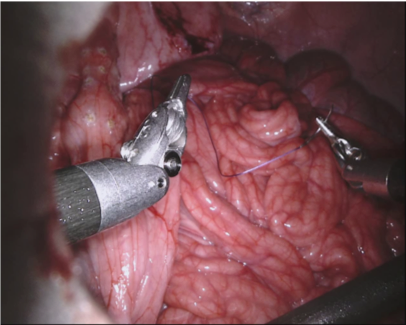}
\caption{Frame 28}
\end{subfigure}
\hfill
\begin{subfigure}[b]{0.23\textwidth}
\includegraphics[width=\textwidth]{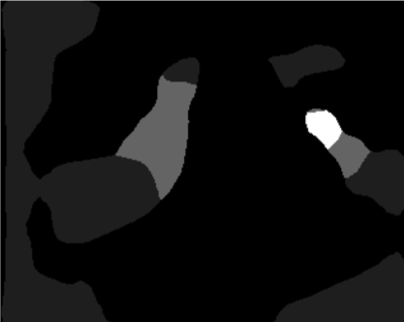}
\caption{U Alberta}
\end{subfigure}
\hfill
\begin{subfigure}[b]{0.23\textwidth}
\includegraphics[width=\textwidth]{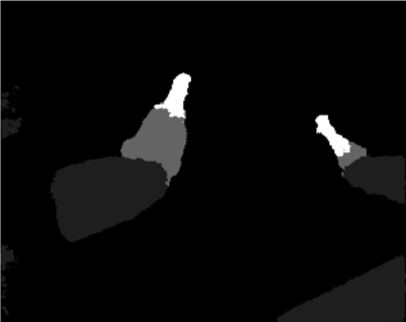}
\caption{SIAT}
\end{subfigure}
\hfill
\begin{subfigure}[b]{0.23\textwidth}
\includegraphics[width=\textwidth]{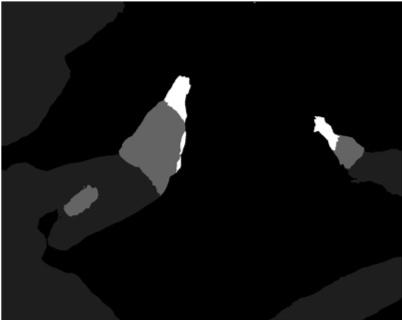}
\caption{UCL}
\end{subfigure}
\caption{\label{fig:parts_segmentation_qualitative} Qualitative results for the parts based labelling showing frames from different datasets with the corresponding results from randomly selected methods. In order, the datasets shown are Dataset 1, Dataset 2, Dataset 5, Dataset 7, Dataset 8 and Dataset 10.}
\end{figure*}

In Fig. \ref{fig:parts_segmentation_qualitative} we show qualitative results from 6 datasets with randomly chosen method outputs for each frame. Again the vessel sealer instrument causes numerous problems with inconsistent labelling occurring all over the shaft. 

\begin{table*}
\centering
\begin{tabular}{ l |  c | c | c | c | c | c | c | c | c | c }
 & NCT & UB & BIT & MIT & SIAT & UCL & TUM & UA & UW & Mean IoU \\
\hline
Dataset 1  & 0.723  & 0.715  & 0.317  & \textbf{0.737}  & 0.591  & 0.611  & 0.708  & 0.485  & 0.235 & 0.569 \\
Dataset 2  & 0.705  & 0.725  & 0.294  & \textbf{0.792}  & 0.632  & 0.606  & 0.740  & 0.559  & 0.244 & 0.589 \\
Dataset 3  & 0.809  & 0.779  & 0.319  & \textbf{0.825}  & 0.753  & 0.692  & 0.787  & 0.640  & 0.239 & 0.649 \\
Dataset 4  & 0.845  & 0.737  & 0.304  & \textbf{0.902}  & 0.792  & 0.630  & 0.815  & 0.692  & 0.238 & 0.662 \\
Dataset 5  & 0.607  & 0.565  & 0.280  & \textbf{0.695}  & 0.509  & 0.541  & 0.624  & 0.473  & 0.240 & 0.504 \\
Dataset 6  & 0.731  & 0.763  & 0.271  & \textbf{0.802}  & 0.677  & 0.668  & 0.756  & 0.608  & 0.235 & 0.612 \\
Dataset 7  & 0.729  & \textbf{0.747}  & 0.359  & 0.655  & 0.604  & 0.523  & 0.727  & 0.438  & 0.207 & 0.554 \\
Dataset 8  & 0.644  & 0.721  & 0.300  & \textbf{0.737}  & 0.496  & 0.441  & 0.680  & 0.604  & 0.236 & 0.540 \\
Dataset 9  & 0.561  & 0.597  & 0.273  & 0.650  & 0.655  & 0.600  & \textbf{0.736}  & 0.551  & 0.221 & 0.538 \\
Dataset 10  & 0.788  & 0.767  & 0.273  & 0.762  & 0.751  & 0.713  & \textbf{0.807}  & 0.637  & 0.241 & 0.638 \\
\hline
Mean IOU & 0.699  & 0.700  & 0.289  & 0.737  & 0.667  & 0.623  & \textbf{0.751}  & 0.578  & 0.357
\end{tabular}
\caption{\label{fig:parts_numerical_results} The numerical results for the parts based segmentation task where the metric used is mean IoU over all classes. The highest scoring method is shown in bold. 7 datasets were won by the team from MIT, 1 by the team from UB and 2 by the team from TUM.}
\end{table*}

\begin{table*}
\centering
\begin{tabular}{ l |  c | c | c | c | c | c | c | c | c | c }
 & NCT & UB & BIT & MIT & SIAT & UCL & TUM & UA & UW & Mean IoU \\
\hline
Dataset 1  & 0.831  & 0.855  & 0.267  & \textbf{0.886}  & 0.634  & 0.692  & 0.820  & 0.402  & 0.241 & 0.625 \\
Dataset 2  & 0.676  & 0.663  & 0.078  & \textbf{0.747}  & 0.614  & 0.579  & 0.703  & 0.543  & 0.256 & 0.540 \\
Dataset 3  & 0.856  & 0.803  & 0.274  & \textbf{0.868}  & 0.794  & 0.724  & 0.809  & 0.688  & 0.272 & 0.676 \\
Dataset 4  & 0.950  & 0.857  & 0.298  & \textbf{0.957}  & 0.913  & 0.817  & 0.923  & 0.829  & 0.611 & 0.795 \\
Dataset 5  & 0.674  & 0.574  & 0.143  & \textbf{0.799}  & 0.712  & 0.663  & 0.759  & 0.560  & 0.259 & 0.571 \\
Dataset 6  & 0.861  & 0.875  & 0.153  & \textbf{0.887}  & 0.769  & 0.789  & 0.861  & 0.755  & 0.325 & 0.697 \\
Dataset 7  & 0.585  & \textbf{0.701}  & 0.154  & 0.388  & 0.453  & 0.312  & 0.640  & 0.148  & 0.075 & 0.384 \\
Dataset 8  & 0.807  & 0.876  & 0.276  & \textbf{0.921}  & 0.450  & 0.420  & 0.788  & 0.861  & 0.349 & 0.639 \\
Dataset 9  & 0.476  & 0.572  & 0.180  & 0.668  & 0.707  & 0.641  & \textbf{0.804}  & 0.652  & 0.318 & 0.558 \\
Dataset 10  & 0.868  & 0.868  & 0.158  & 0.859  & 0.867  & 0.806  & \textbf{0.900}  & 0.778  & 0.559 & 0.740 \\
\hline
Mean IOU & 0.727  & 0.751  & 0.188  & 0.786  & 0.729  & 0.676  & 0.822  & 0.660 & 0.373
\end{tabular}
\caption{\label{fig:parts_shaft_numerical_results} The numerical results for the shaft component of parts based segmentation task where the metric used is mean IoU over all classes. The highest scoring method is shown in bold. 7 datasets were won by the team from MIT, 1 by the team from UB and 2 by the team from TUM.}
\end{table*}

\begin{table*}
\centering
\begin{tabular}{ l |  c | c | c | c | c | c | c | c | c | c }
 & NCT & UB & BIT & MIT & SIAT & UCL & TUM & UA & UW & Mean IoU \\
\hline
Dataset 1  & \textbf{0.581}  & 0.522  & 0.009  & 0.530  & 0.399  & 0.438  & 0.567  & 0.318  & 0.089 & 0.383 \\
Dataset 2  & 0.592  & 0.598  & 0.014  & \textbf{0.681} & 0.465  & 0.404  & 0.611  & 0.374  & 0.062 & 0.422 \\
Dataset 3  & 0.756  & 0.730  & 0.012  & \textbf{0.766}  & 0.665  & 0.566  & 0.728  & 0.561  & 0.140 & 0.547 \\
Dataset 4  & 0.656  & 0.382  & 0.004  & \textbf{0.765}  & 0.540  & 0.232  & 0.571  & 0.381  & 0.000 & 0.392 \\
Dataset 5  & 0.457  & 0.440  & 0.005  & \textbf{0.565}  & 0.246  & 0.397  & 0.390  & 0.282  & 0.027 & 0.312 \\
Dataset 6  & 0.594  & 0.687  & 0.011  & \textbf{0.727}  & 0.572  & 0.551  & 0.676  & 0.481  & 0.123 & 0.491 \\
Dataset 7  & \textbf{0.691}  & 0.627  & 0.012  & 0.654  & 0.380  & 0.428  & 0.588  & 0.352  & 0.085 & 0.424 \\
Dataset 8  & 0.665  & 0.686  & 0.019  & 0.720  & 0.435  & 0.333  & \textbf{0.774}  & 0.501  & 0.023 & 0.462 \\
Dataset 9  & 0.354  & 0.371  & 0.000  & 0.475  & 0.407  & 0.372  & \textbf{0.562}  & 0.255  & 0.013 & 0.312 \\
Dataset 10  & 0.711  & 0.664  & 0.000  & 0.672  & 0.641  & 0.583  & \textbf{0.742}  & 0.512  & 0.145 & 0.519 \\
\hline
Mean IOU & 0.578  & 0.551  & 0.005  & 0.625  & 0.495 & 0.449  & \textbf{0.634}  & 0.396  & 0.074 & 
\end{tabular}
\caption{\label{fig:parts_wrist_numerical_results} The numerical results for the wrist component of parts based segmentation task where the metric used is mean IoU over all classes. The highest scoring method is shown in bold. 5 datasets were won by the team from MIT, 2 by the team from NCT and 3 were won by the team from TUM.}
\end{table*}

\begin{table*}
\centering
\begin{tabular}{ l |  c | c | c | c | c | c | c | c | c | c  }
 & NCT & UB & BIT & MIT & SIAT & UCL & TUM & UA & UW & Mean IoU \\
\hline
Dataset 1  & 0.491  & 0.494  & 0.058  & \textbf{0.538}  & 0.356  & 0.337  & 0.460  & 0.272  & 0.082 & 0.343 \\
Dataset 2  & 0.588  & 0.656  & 0.106  & \textbf{0.743}  & 0.501  & 0.485  & 0.665  & 0.380  & 0.000 & 0.458 \\
Dataset 3  & 0.634  & 0.593  & 0.070  & \textbf{0.676}  & 0.565  & 0.497  & 0.620  & 0.337  & 0.000 & 0.444 \\
Dataset 4  & 0.782  & 0.720  & 0.031  & \textbf{0.892}  & 0.727  & 0.490  & 0.776  & 0.588  & 0.175 & 0.576 \\
Dataset 5  & 0.332  & 0.260  & 0.005  & \textbf{0.433}  & 0.131  & 0.131  & 0.381  & 0.085  & 0.000 & 0.195 \\
Dataset 6  & 0.483  & 0.503  & 0.026  & \textbf{0.606}  & 0.385  & 0.354  & 0.501  & 0.225  & 0.052 & 0.348 \\
Dataset 7  & \textbf{0.613}  & 0.582  & 0.274  & 0.494  & 0.458  & 0.271  & \textbf{0.613}  & 0.283  & 0.313 & 0.433 \\
Dataset 8  & 0.111  & 0.331  & 0.002  & \textbf{0.314}  & 0.151  & 0.067  & 0.165  & 0.061  & 0.001 & 0.134 \\
Dataset 9  & 0.421  & 0.449  & 0.000  & 0.462  & 0.512  & 0.398  & \textbf{0.583}  & 0.318  & 0.033 & 0.353 \\
Dataset 10  & 0.583  & 0.543  & 0.000  & 0.529  & 0.504  & 0.475  & \textbf{0.596}  & 0.272  & 0.057 & 0.395 \\
\hline
Mean IOU & 0.503  & 0.507  & 0.034  & 0.542  & 0.460  & 0.384  & \textbf{0.556}  & 0.288  & 0.060 &
\end{tabular}
\caption{\label{fig:parts_clasper_numerical_results} The numerical results for the clasper component of parts based segmentation task where the metric used is mean IoU over all classes. The highest scoring method is shown in bold. 7 datasets were won by the team from MIT, 1 was tied between the team from NCT and the team from TUM and 2 were won outright by the team from TUM.}
\end{table*}

\subsection{Type Segmentation}

The final challenge was to identify each instrument type from the list of Large Needle Driver, Prograsp Forceps, Monopolar Curved Scissors, Vessel Sealer, Fenestrated Bipolar Forceps and Grasping Retractor (see Fig. \ref{fig:inst_types}). Only 6 teams participated in this challenge, due particularly to the significant increase in difficulty in recognizing many of the da Vinci instruments from one another.  


\begin{table*}
\centering
\begin{tabular}{ l |  c | c | c | c | c | c | c }
 & NCT & UB & MIT & SIAT & UCL & UA & Mean IoU \\
\hline
Dataset 1  & 0.056  & 0.111  & \textbf{0.177}  & 0.138  & 0.073  & 0.068 & 0.104 \\
Dataset 2  & 0.499  & 0.722  & \textbf{0.766}  & 0.013  & 0.481  & 0.244 & 0.454 \\
Dataset 3  & \textbf{0.926}  & 0.864  & 0.611  & 0.537  & 0.496  & 0.765 & 0.690 \\
Dataset 4  & 0.551  & 0.680  & \textbf{0.871}  & 0.223  & 0.204  & 0.677 & 0.534 \\
Dataset 5  & 0.442  & 0.443  & \textbf{0.649}  & 0.017  & 0.301  & 0.001 & 0.309 \\
Dataset 6  & 0.109  & 0.371  & \textbf{0.593}  & 0.462  & 0.246  & 0.400 & 0.363 \\
Dataset 7  & 0.393  & \textbf{0.416}  & 0.305  & 0.102  & 0.071  & 0.000 & 0.215 \\
Dataset 8  & 0.441  & 0.384  & \textbf{0.833}  & 0.028  & 0.109  & 0.357 & 0.359 \\
Dataset 9  & 0.247  & 0.106  & \textbf{0.357}  & 0.315  & 0.272  & 0.040 & 0.223 \\
Dataset 10  & 0.552  & 0.709  & 0.609  & \textbf{0.791}  & 0.583  & 0.715 & 0.660 \\
\hline
Mean IOU & 0.409  & 0.453  & \textbf{0.542}  & 0.371  & 0.337  & 0.346 &
\end{tabular}
\caption{\label{fig:type_numerical_results} The numerical results for the type based segmentation task where the metric used is mean IoU over all classes. The highest scoring method is shown in bold. 7 datasets were won by the team from MIT, 1 was won by the team from SIAT, 1 by the team from NCT and 1 by the team from UB.}
\end{table*}

\begin{figure*}
\captionsetup[subfigure]{labelformat=empty}
\begin{subfigure}[b]{0.19\textwidth}
\includegraphics[width=\textwidth]{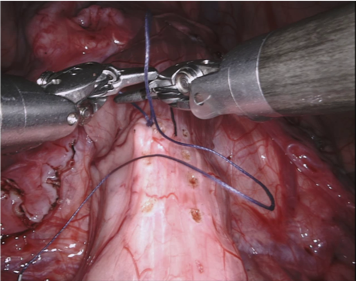}
\caption{Frame 229}
\end{subfigure}
\hfill
\begin{subfigure}[b]{0.19\textwidth}
\includegraphics[width=\textwidth]{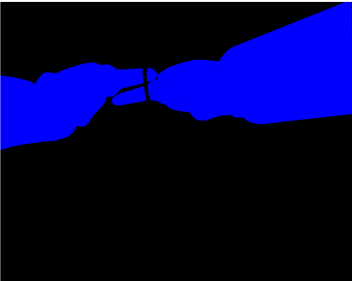}
\caption{Ground Truth}
\end{subfigure}
\begin{subfigure}[b]{0.19\textwidth}
\includegraphics[width=\textwidth]{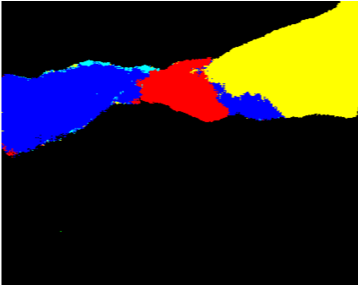}
\caption{SIAT}
\end{subfigure}
\hfill
\begin{subfigure}[b]{0.19\textwidth}
\includegraphics[width=\textwidth]{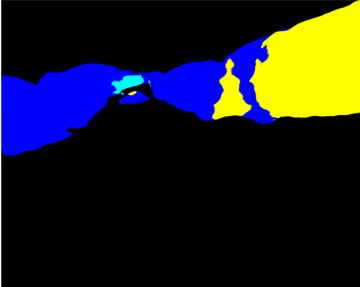}
\caption{MIT}
\end{subfigure}
\hfill
\begin{subfigure}[b]{0.19\textwidth}
\includegraphics[width=\textwidth]{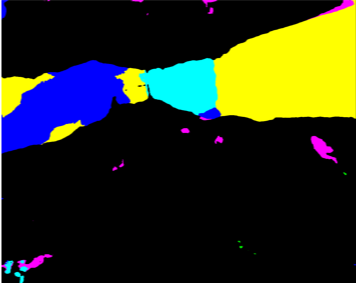}
\caption{UCL}
\end{subfigure}
\\
\begin{subfigure}[b]{0.19\textwidth}
\includegraphics[width=\textwidth]{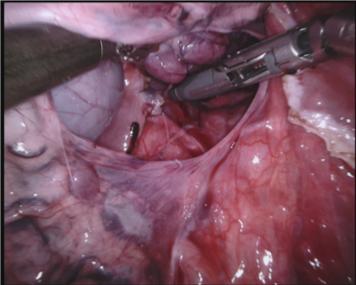}
\caption{Frame 240}
\end{subfigure}
\hfill
\begin{subfigure}[b]{0.19\textwidth}
\includegraphics[width=\textwidth]{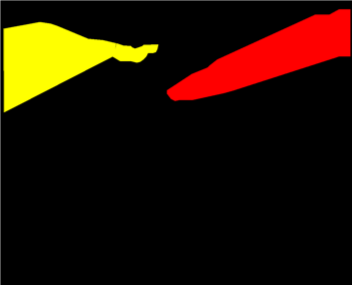}
\caption{Ground Truth}
\end{subfigure}
\begin{subfigure}[b]{0.19\textwidth}
\includegraphics[width=\textwidth]{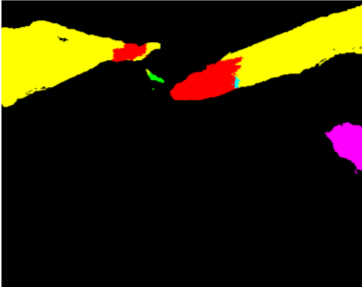}
\caption{U Bern}
\end{subfigure}
\hfill
\begin{subfigure}[b]{0.19\textwidth}
\includegraphics[width=\textwidth]{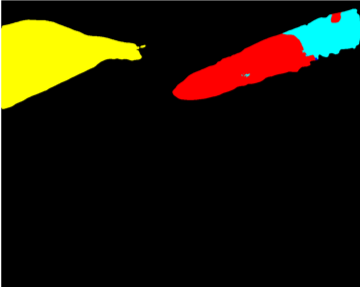}
\caption{MIT}
\end{subfigure}
\hfill
\begin{subfigure}[b]{0.19\textwidth}
\includegraphics[width=\textwidth]{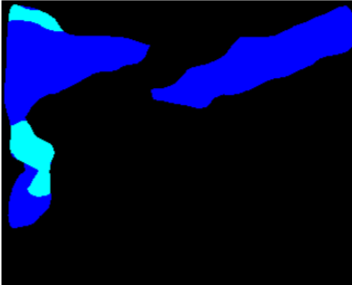}
\caption{U Alberta}
\end{subfigure}
\\
\begin{subfigure}[b]{0.19\textwidth}
\includegraphics[width=\textwidth]{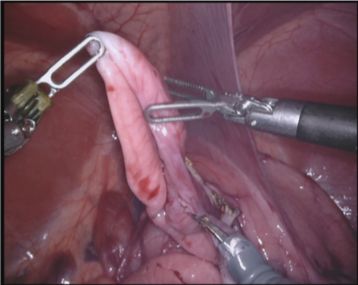}
\caption{Frame 59}
\end{subfigure}
\hfill
\begin{subfigure}[b]{0.19\textwidth}
\includegraphics[width=\textwidth]{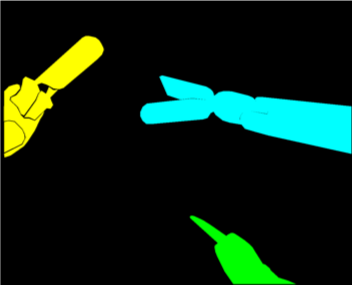}
\caption{Ground Truth}
\end{subfigure}
\begin{subfigure}[b]{0.19\textwidth}
\includegraphics[width=\textwidth]{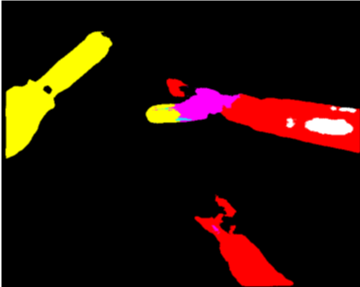}
\caption{NCT}
\end{subfigure}
\hfill
\begin{subfigure}[b]{0.19\textwidth}
\includegraphics[width=\textwidth]{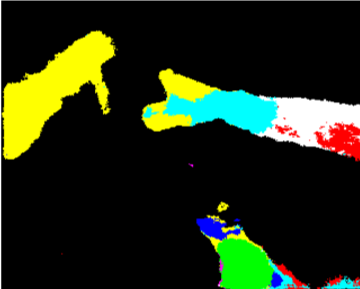}
\caption{SIAT}
\end{subfigure}
\hfill
\begin{subfigure}[b]{0.19\textwidth}
\includegraphics[width=\textwidth]{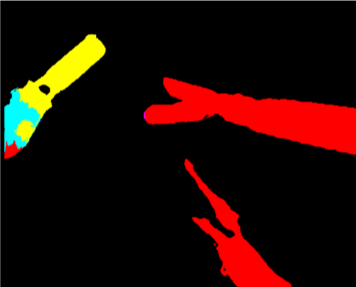}
\caption{U Bern}
\end{subfigure}
\\
\caption{\label{fig:type_segmentation_qualitative} Quantitative Results for the type based segemtation. The ground truth frame shows the correct labelling for each pixel. In order, the datasets shown are Dataset 3, 5 and 9. Although the masks are quite accurate for many of the frames, identification is much poorer with many instruments either partially or completely misclassified.}
\end{figure*}

\section{Discussion and conclusions}

Different methods have been proposed to tackle the problem of creating CNNs designed to be an efficient architecture for pixel-wise semantic segmentation. These networks can produce a segmentation map for an entire input image in a single forward pass. One of the most successful state-of-the-art deep learning methods is based on the FCN \cite{long_fully_2015}.  The main idea of this approach is to use CNN as a powerful feature extractor by replacing the fully connected layers with convolutional layers to output spatial feature maps instead of classification scores. Those maps are further up-sampled to produce dense pixel-wise output. This method has been further improved and is now known as U-Net neural network \cite{ronneberger_unet_2015}. The U-Net architecture uses skip connections to combine low-level feature maps with higher-level ones, which enables precise pixel-level localization. A large number of feature channels in the up-sampling part allows propagating context information to higher resolution layers. To date, the networks based on U-Net have become standard tools in segmentation tasks for medical images. Further improvements take into account encoders such as VGG, ResNets etc. pre-trained on ImageNet \cite{iglovikov2018ternausnet}. 

\subsection{Data Augmentation}

Data augmentation is a common technique used in deep learning to increase the size and variability of the training dataset by making geometric or photometric modifications such as rotations, croppings, and modifications to the color palette. Given the correlated nature of the data in a single sequence, data augmentation provides a simple and effective way of improving generalization and reducing overfitting. Endoscopic surgical robotic data differs vastly from computer vision tasks on natural images, making it more challenging to find data augmentation techniques which are advantageous for the model. Certain techniques may even degenerate the performance as model capacity is used for scenes which never appears in the test set.

\begin{table*}
\centering
\begin{tabular}{l|c|c|c|c|c|c|c|c|c|c}
Augmentation & NCT & UB & BIT & MIT & SIAT & UCL & TUM & IIT & UA & UW \\
\hline
Horizontal Flips & \checkmark & \checkmark & ? & \checkmark & \checkmark &  & \checkmark & \checkmark & &  \\
Vertical Flips & \checkmark & \checkmark & ? & \checkmark & \checkmark &  & & & &  \\
Zooms/Crops & \checkmark & & ? & & & & \checkmark & & &  \\
Translations & & \checkmark & ? & & & & \checkmark & & &  \\
Rotations & \checkmark & \checkmark & ? & & & & \checkmark & & &  \\
Color & \checkmark & & ? & \checkmark & & & \checkmark & & &  \\
Normalization & & \checkmark & ? & \checkmark & & & \checkmark & & &  \\
Contrast & & & ? & & & & \checkmark & & &  \\
Specular Highlights & & & ? & & & & \checkmark & & &  \\
\end{tabular}
\caption{\label{fig:table_augmentations}The different data augmentation selections made by the participating teams. The column for University of Washington (UW) is blank as they pursued a non-deep learning approach and therefore did not perform any data augmentation. The columns for University of Alberta (UA) and UCL are also blank as they did not perform data augmentation due to time limitations. We currently have no data for the augmentations used by the team from Beijing Institute of Technology (BIT).}
\end{table*}

Most teams performed spatial augmentations rather than photometric augmentations which makes sense given that all of the images in the training set were captured with the da Vinci Xi endoscopic camera. This is a factory calibrated camera with a limited set of color display settings and a single Xenon light source illuminating the scene which leads to fairly uniform images. Typically the variation of observed scene orientations in robot surgery is larger than real world images which are usually taken with cameras in a upright orientation. The most common spatial augmentations were to flip the images horizontally and vertically which fits with the set of common orientations that are observed in robotic surgery. The physical constraints created by the robot design allow instruments to typically enter the field of view from the side of the image and point towards a single central point due to the triangulation created by the design of the arms. The limited use of zooming transformations, either as crops or zoom outs with padding was surprising given that this is the most common type of image shift observed during surgery. 

Table \ref{fig:table_augmentations} illustrates the augmentation choices made by the participating teams. All of the highest performing teams conducted some photometric augmentation alongside the ubiquitous spatial augmentations suggesting that these are still important to obtain good performance. One particularity in endoscopic surgery data is due to the directed light source that results in specular reflections on the instrument as well as the tissue. This poses an additional difficulty to segmentation algorithms. To address this problem, the team of the Technical University of Munich proposed to augment the training data by placing simple specular reflections randomly in the image which improved the segmentation result. 


\begin{figure}
\begin{subfigure}[b]{0.31\columnwidth}
\includegraphics[width=\textwidth]{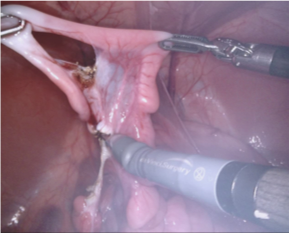}
\caption{}
\end{subfigure}
\hfill
\begin{subfigure}[b]{0.31\columnwidth}
\includegraphics[width=\textwidth]{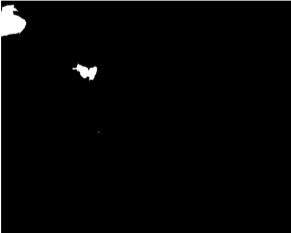}
\caption{}
\end{subfigure}
\hfill
\begin{subfigure}[b]{0.31\columnwidth}
\includegraphics[width=\textwidth]{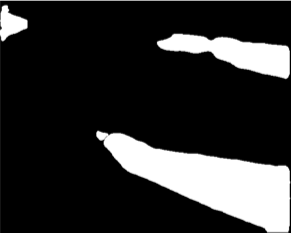}
\caption{}
\end{subfigure}
\begin{subfigure}[b]{0.31\columnwidth}
\includegraphics[width=\textwidth]{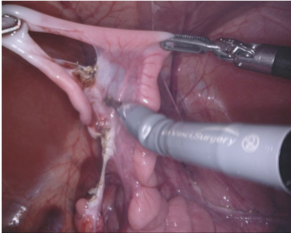}
\caption{}
\end{subfigure}
\hfill
\begin{subfigure}[b]{0.31\columnwidth}
\includegraphics[width=\textwidth]{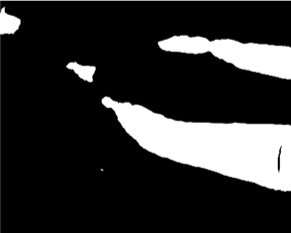}
\caption{}
\end{subfigure}
\hfill
\begin{subfigure}[b]{0.31\columnwidth}
\includegraphics[width=\textwidth]{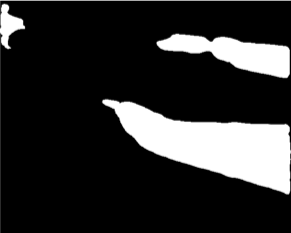}
\caption{}
\end{subfigure}
\caption{\label{fig:fails_smoke} Interesting failure case in a mildly smoky scene where the instrument appearance is clearly observable to the human eye. (a) shows a frame with a thin smoke layer from usage of the Monopolar Curved Scissors. (b) shows the binary segmentation of this frame by the UB method and (c) shows the binary segmentation from the method of TUM. (d-f) shows the same setup for the next frame where the smoke has cleared. Despite UB and TUM having very similar overall accuracy, the method of UB has near total failure when the smoke is present.}
\end{figure}
\subsection{Challenge Design}

There are several limitations with the challenge design. The largest and most significant limitation is the relatively small size of the dataset, being made up of only 3000 frames of which 1800 were selected as training data. This is much smaller in comparison to industry standard computer vision segmentation datasets such as Microsoft COCO \cite{lin_coco_2015} (328k images). The selection of data from a small number of procedures is typical in surgical datasets due to difficulty of obtaining data from varied sources however, it creates bias in the ranking towards models which can overfit without properly testing generalization. Moving forwards video segmentation challenges should aim to reduce the number of frames contributed by each dataset while increasing the number of separate procedures. A related issue which occurs due to sampling from a small number of procedure videos is that the occurrence of different instrument types in the training dataset may not accurately reflect their true distribution which is set by by how often the instrument is used in surgery. For instance, the Monopolar Curved Scissors appeared in just 2 training sequences yet is a commonly used instrument in da Vinci procedures. 

\begin{figure}
\includegraphics[width=0.48\textwidth]{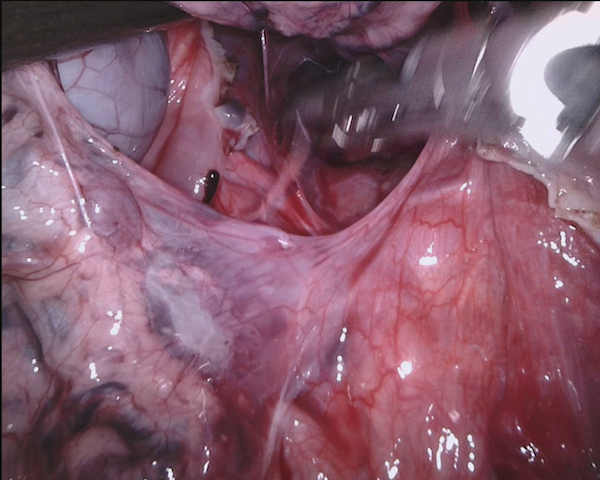}
\caption{\label{fig:ambig_blur} An example of ambiguous labelling needs for a blurred frame where due to instrument motion it is not clear where the border should lie.}
\end{figure}

Labelling errors also create complications for the challenge participants. Alongside standard errors due to carelessness amongst annotators, fast instrument motion causing image blur (see Fig. \ref{fig:ambig_blur}) may result in labelling errors. Severe ambiguities which confuse even a human annotator due to lighting or smoke which can be common in laparoscopic procedures are not present in our datasets. Recent work \cite{maier_hein_can_2014} on reducing  errors by validating all annotations with repetitions and majority voting is an effective strategy although it requires a significant duplication of effort. A simpler and faster alternative may be to require different annotators verify each sample \cite{lin_coco_2015}. Due to delays in beginning the annotation and a limited time window to release the data before the challenge deadline, we only had sufficient time for a single trained annotator to complete each sample with review by a single expert. Limitations in the labelling software used to label the images also creates inconsistencies. As many of the da Vinci grasping instruments contain one or more holes in the clasper and these should be correctly reflected in the annotations. However, the annotation tool used in this challenge did not allow holes in the polygons and our training instructions did not explicitly request that the annotators spent the additional time to ensure that the hole was properly annotated. By the time this problem was identified there was not sufficient time to correct the annotations before the challenge data release and to ensure consistency across training and test set, the hole was not reannotated in any frames. 

A further limitation of the challenge design is that it expected teams to make submissions for 10 datasets requiring 9 separate models to be trained. Although this increased the number of sequences that had associated test data, many teams felt that the computational requirement of training and evaluating so many models was too high to allow for proper experimentation. In future challenges, we plan to release entirely separate training and test sets so that only a single model needs to be evaluated.

\subsection{Future Challenges}

Although segmentation of robotic instruments is an interesting and important problem, there is significant value to providing dense segmentation entire scene by annotating the anatomy as well as the instruments. Algorithms with the ability to recognize different tissues would have the potential to provide much more context aware assistance to surgeons. In 2018 we released a further dataset\footnote{\url{https://endovissub2018-roboticscenesegmentation.grand-challenge.org/}} with complete annotation of several different tissue types as well as multiple third party devices such as clips, thread and suction/irrigation tools.

\bibliographystyle{IEEEtran}
\bibliography{lib}

\end{document}